\pdfoutput=1
\documentclass[conference]{IEEEtran}
\usepackage{amsmath,amssymb,amsfonts}
\usepackage{algorithm}
\usepackage{algpseudocode}
\usepackage{graphicx}
\usepackage{booktabs}
\usepackage{threeparttable}
\usepackage{xcolor}
\usepackage{cite}
\usepackage[hidelinks]{hyperref}
\newcommand{\R}{\mathbb{R}}
\newcommand{\eFCM}{e-FCM}
\graphicspath{{figures/}}

\begin{document}

\title{Fuzzy e-Means: Fuzzy c-Means on Large Data\\ with Moving-Average Mini-Batch Updates}

\author{\IEEEauthorblockN{Paritosh Tiwari\IEEEauthorrefmark{1} and Punit Rathore\IEEEauthorrefmark{1}\IEEEauthorrefmark{2}}
\IEEEauthorblockA{\IEEEauthorrefmark{1}Robert Bosch Centre for Cyber-Physical Systems (RBCCPS)\\
\IEEEauthorrefmark{2}Centre for Infrastructure, Sustainable Transportation and Urban Planning (CiSTUP)\\
Indian Institute of Science, Bengaluru, India. Email: paritosht@iisc.ac.in}}

\maketitle

\begin{abstract}
Fuzzy c-means (FCM) gives every point a degree of membership in every cluster. Its standard solver, alternating
optimization (AO), needs a full pass over the data per iteration and often many iterations, while methods built for
large data stop at a worse objective or still need tens of passes. We run FCM with the optimizer of e-means, a recent
mini-batch k-means that keeps moving averages of the number and the sum of the points in each cluster and sets its
batch size and step size by scaling laws. The optimizer carries over because an FCM centroid is also a weighted sum
divided by a total weight, but this direct transfer fails in two ways: the e-means rule that relocates nearly empty
clusters also removes the centroids of small, distant groups, and FCM started from random points can settle in poor
local minima. Fuzzy e-means (e-FCM) therefore starts with one pass of e-means with hard assignments and then runs fuzzy
moving-average passes without relocation. On 28 configurations with up to 109.5 million points and 16,384 clusters,
e-FCM reaches the objective of 25 AO iterations in two to four passes when clusters hold several hundred points or
more, 11 to 20 times faster than AO on one-million-point embedding sets with 1,024 clusters. It ends within 0.16\% of
converged AO on these sets and below converged AO where AO from random points lands in poor minima, but it is slower
than AO on small low-dimensional problems, where an AO pass takes milliseconds. Finally, the common fuzzifier $m=2$
makes the centroids collapse or merge on most high-dimensional sets, and e-FCM can choose the fuzzifier by matching a
known overlap at about two passes per candidate value.
\end{abstract}


\section{Introduction}
\label{sec:intro}
Fuzzy c-means (FCM) \cite{dunn1973,bezdek1981} is the soft counterpart of k-means. Instead of assigning each point to
one cluster, it gives the point a membership in every cluster, a number between 0 and 1 that decreases with the
distance to the cluster's centroid; the memberships of a point sum to 1. This suits data whose groups overlap: a voxel
of a brain scan that straddles grey and white matter contains both tissues, and a satellite pixel can cover both forest
and field. FCM is usually solved by alternating optimization (AO), which computes the memberships of all points for
fixed centroids, moves every centroid to the membership-weighted mean of the data, and repeats. Each AO iteration is
therefore a full pass over the data, and AO often needs many of them. On one million 1024-dimensional image embeddings
with 16,384 clusters, for example, one AO iteration takes 1.4\,s on our GPU, the 200 iterations we ran take 285\,s,
and after 25 iterations the objective is still 0.44\% above its final value.

Methods for running FCM on large data save passes in three ways \cite{havens2012}. Sampling methods cluster a random
sample and extend the result to all points \cite{havens2012,cheng1998fast}, incremental methods cluster the data chunk
by chunk and combine the results \cite{hore2007spfcm,hore2008ofcm}, and stochastic methods update the centroids after
every point or mini-batch \cite{wang2014sgfc}. Each family has a weakness. When there are many clusters, a sample or a
chunk holds few points per cluster, so the centroids it yields are noisy. A stochastic update, on the other hand,
usually moves a centroid by a step that shrinks as more points are assigned to its cluster, as in the online k-means
of MacQueen and the mini-batch k-means of Sculley \cite{macqueen1967,sculley2010}. Early updates, made while the
centroids were still poor, then keep their weight, and the centroids slow down before they reach a minimum.

For k-means, a recent method, e-means \cite{bessa2026emeans}, addresses this weakness of the stochastic approach. For
every cluster, it keeps an exponential moving average of the number of points assigned to it, which we call the
cluster's mass, and of their sum, so that the statistics of old batches fade instead of piling up; the centroid is the
ratio of the two. Scaling laws, fitted once on k-means problems, set the batch size and the step size of the moving
average, and a rule called revival relocates clusters that receive almost no points. Without any tuning, e-means
reaches the loss of 25 iterations of Lloyd's k-means algorithm in about two passes. An FCM centroid has the same form
as the centroid that e-means maintains: it is a weighted sum of points divided by the sum of the weights, with
membership weights in place of the 0/1 assignments of k-means. We therefore ask whether the e-means optimizer, with
counts and sums replaced by their membership-weighted versions, makes FCM converge faster, and how it compares with
the other ways of running FCM on large data.

We call this replacement the direct transfer of e-means to FCM. It is fast on balanced data, but it fails in two ways.
The first concerns revival, the part of e-means with the largest effect on k-means \cite{bessa2026emeans}: a cluster
whose mass falls below a tenth of the average is treated as dead and moved onto a large cluster. Under FCM, the mass of
a cluster is the sum of its membership weights, and a cluster that covers a small, distant group of points also has a
small mass. Revival then moves the centroid away from the group it covers. On the KDD Cup 99 network-connection
data, the direct transfer ends 29 to 68\% above the objective of AO run to convergence for this reason. The second
failure concerns the start: FCM started from randomly chosen points sometimes settles in a poor local minimum, and a
mini-batch method keeps the solution that its first pass picks. Fuzzy e-means (\eFCM) avoids both failures with one
change. It first runs one epoch, that is, one pass over the data, of e-means with hard assignments, which is cheap, places
the centroids in a k-means solution and relocates dead clusters at a stage where a small count does mean an unused
cluster. \eFCM\ then runs fuzzy moving-average epochs without revival. We call the two stages the hard
phase and the fuzzy phase.

We compare \eFCM\ with AO and six other ways of running FCM on large data, all on one GPU with the same membership
code, the same starting centroids and the same exact objective. A standard benchmark holds seven datasets used in
earlier studies of FCM for large data and of e-means, with up to one million points and 16,384 clusters. These data
raise a problem of their own. FCM has a parameter, the fuzzifier $m>1$, that sets how soft the memberships are. With
the common choice $m=2$, all centroids can collapse onto the mean of the data on five of the seven datasets, and most
centroids merge on a sixth; on the embedding sets, the usable values are so close to 1 that FCM behaves almost like
k-means. A second, fuzzy benchmark therefore uses data whose groups overlap: Gaussian
mixtures with a known overlap, a simulated brain MRI scan with the true tissue fraction of every voxel, and 109.5
million satellite pixels.

The answer to our question has three parts. First, the saving in passes carries over from k-means: \eFCM\ reaches the
objective that AO reaches after 25 iterations in two to four passes when clusters hold several hundred points or more,
and in a median of two passes on 13 of the 14 configurations of the fuzzy benchmark. Second, fewer passes mean less
time only
where a pass is not already cheap: \eFCM\ is 11 to 20 times faster than AO on the embedding sets with 1,024 clusters,
but slower on small three-dimensional problems. Third, \eFCM\ ends close to AO run to convergence, and below it where
AO started from random points lands in poor minima, whereas mini-batch FCM stalls higher and the chunk- and
sample-based methods need tens of passes. Of the parts of e-means, the moving average is essential and the scaling
laws carry over without refitting on the standard benchmark, but revival has to be removed from the fuzzy phase.

Our contributions are the following.
\begin{itemize}
\item \eFCM, which runs FCM with the e-means optimizer and moves the relocation of centroids into one hard epoch
(Section~\ref{sec:method}).
\item An analysis of why relocation by mass fails under FCM: a small mass marks a small, distant group as well as an
unused cluster, and triggering relocation on hard counts instead reduces the damage but does not remove it
(Section~\ref{sec:components}).
\item A GPU comparison of the ways of running FCM on large data on 28 configurations, with the same membership code,
initialization and exact objective for all methods (Sections~\ref{sec:setup} and~\ref{sec:results}).
\item Findings on the fuzzifier at scale: $m=2$ makes the centroids collapse or merge on most high-dimensional sets;
matching a known overlap gives $m=1.3$ to $1.8$ in three dimensions but values close to 1 once noise dimensions are
added; and \eFCM\ can carry out this matching at about two passes per candidate value (Section~\ref{sec:fuzzifier}).
\end{itemize}

\section{Background}
\label{sec:background}

\subsection{Fuzzy c-Means and Its Reduced Objective}
Let $x_1,\dots,x_n\in\R^d$ be the points and $C=(c_1,\dots,c_k)$ the centroids, and write $D_{ij}=\|x_i-c_j\|^2$. FCM
\cite{dunn1973,bezdek1981} minimizes $J_m(U,C)=\sum_{i,j}u_{ij}^m D_{ij}$ over memberships $u_{ij}\ge0$ with
$\sum_j u_{ij}=1$ for every point. The fuzzifier $m>1$ sets how soft the memberships are: as $m\to1$ they become the
0/1 assignments of k-means, and as $m$ grows they tend to the uniform value $1/k$. For fixed centroids the optimal
memberships are
\begin{equation}
u_{ij}=\frac{D_{ij}^{-1/(m-1)}}{\sum_{l}D_{il}^{-1/(m-1)}}.
\label{eq:u}
\end{equation}
Substituting them into $J_m$ gives the reduced objective \cite{hathaway1995}
\begin{equation}
R_m(C)=\frac1n\sum_{i=1}^n\Big(\sum_{j=1}^k D_{ij}^{-1/(m-1)}\Big)^{1-m}=\frac1n\min_U J_m(U,C),
\label{eq:R}
\end{equation}
which depends on the centroids alone. It therefore scores the output of every method on the same footing, and we use
it throughout: on the training points, whose objective all methods minimize, and on held-out points, which show
whether a solution generalizes. As $m\to1$, $R_m$ tends to the mean squared error of k-means.

The standard solver, AO for FCM (AO-FCM) \cite{bezdek1981}, alternates between \eqref{eq:u} and the centroid update
$c_j=S_j/N_j$, with $S_j=\sum_i u_{ij}^m x_i$ and $N_j=\sum_i u_{ij}^m$: each centroid moves to the membership-weighted
mean of the data. Each iteration decreases $J_m$ \cite{bezdek1987conv,groll2005} and is one pass over the data. A pass
costs $O(nkd)$, the same order as an iteration of Lloyd's algorithm, but it needs all $k$ memberships of every point
instead of one nearest centroid.

Large fuzzifiers make FCM degenerate. The mean of the data, with every centroid on it, is a fixed point of AO-FCM, and
it becomes a stable one once $m>m_G=1/(1-2\lambda_{\max})$, where $\lambda_{\max}$ is the largest eigenvalue of the
data covariance divided by its trace \cite{yu2004}. We call $m_G$ the collapse ceiling; there is no such ceiling when
$\lambda_{\max}\ge1/2$. Below $m_G$, groups of centroids can still merge \cite{winkler2011,schwammle2010}.
High-dimensional data spread their variance over many directions, so $\lambda_{\max}$ is small and $m_G$ lies close
to 1.

\subsection{FCM for Large Data}
The methods we compare with belong to the three families of Section~\ref{sec:intro} \cite{havens2012}; speed-ups that
aggregate similar points \cite{eschrich2003} or make one iteration cheaper \cite{kolen2002} are orthogonal to them.
Among the sampling methods, random sample and extend FCM (rseFCM) \cite{havens2012} clusters one sample, and
multistage random sampling runs FCM on growing samples and uses the result to start FCM on the full data
\cite{cheng1998fast}. Among the incremental methods, single-pass FCM (spFCM) \cite{hore2007spfcm} clusters each chunk
together with the centroids of the previous chunks, which carry weights $\sum_i u_{ij}$, and online FCM (oFCM)
\cite{hore2008ofcm} clusters each chunk separately and then clusters all chunk centroids with their weights; both run
weighted FCM to convergence on each chunk. Among the stochastic methods, stochastic-gradient FCM \cite{wang2014sgfc}
takes the steps of MacQueen's and Sculley's k-means \cite{macqueen1967,sculley2010,bottou1994}: the step of each
centroid is the weight that the batch gives its cluster, divided by the total weight that the cluster has received so
far. Recursive FCM adds a forgetting factor for processes that drift over time \cite{dovzan2011}.

\subsection{e-Means}
e-means \cite{bessa2026emeans} is a mini-batch k-means with five parts. An epoch is one pass over the data in
shuffled batches.

\emph{Moving-average statistics.} Each cluster keeps a mass $N_j$, a moving count of its points, and a sum $S_j$, and
its centroid is $c_j=S_j/N_j$. After each batch, both are updated by an exponential moving average (EMA),
\begin{equation}
N_j\leftarrow(1-\alpha_t)N_j+\alpha_t n_j,\qquad S_j\leftarrow(1-\alpha_t)S_j+\alpha_t s_j,
\label{eq:ema}
\end{equation}
where $n_j$ and $s_j$ are the number and the sum of the batch points whose nearest centroid is $c_j$
\cite{oord2017,neal1998}. With cumulative statistics, the weight of a new batch shrinks as its cluster grows; with the
EMA it stays of the order of $\alpha_t$, so a centroid can still move when its assignments change.

\emph{Cosine schedule.} The step $\alpha_t$ falls from its peak $\alpha_0$ to $\alpha_0/10$ along a cosine over the
run.

\emph{Scaling laws.} The batch size $B$ is a power law in the number of points $n$, of clusters $k$ and of epochs $T$,
and $\alpha_0$ a power law in $n$, $k$, $B$ and $T$. The e-means authors fitted the laws once on k-means runs, so no
tuning is done. The implementation in the Faiss library \cite{douze2024faiss}, which we use, switches to a law with
eight times smaller batches when the dimension $d$ is 16 or less.

\emph{Revival.} Every five batches, each cluster whose mass has fallen below a tenth of the mean mass takes half of a
donor cluster, drawn in proportion to the masses, whose centroid is split in two.

\emph{Closing.} The returned centroids are the means of the points assigned during the last epoch.

\section{Fuzzy e-Means}
\label{sec:method}
\eFCM\ keeps the e-means optimizer and changes two things: the statistics it averages, which become soft, and the place
where it relocates centroids, which becomes a short hard phase before the fuzzy one.

\subsection{Soft Statistics: The Direct Transfer}
The FCM centroid $c_j=S_j/N_j$ has exactly the form that e-means maintains, with soft weights $u_{ij}^m$ in place of
hard 0/1 assignments. The direct transfer of e-means to FCM therefore computes, for every batch $\mathcal{B}$, the
memberships \eqref{eq:u} of its points against the current centroids, and replaces the count and the sum of each
cluster by its soft mass and soft sum,
\begin{equation}
n_j=\sum_{i\in\mathcal{B}}u_{ij}^m,\qquad s_j=\sum_{i\in\mathcal{B}}u_{ij}^m x_i.
\label{eq:soft}
\end{equation}
Everything else is kept: the EMA \eqref{eq:ema}, the cosine schedule, the scaling laws (not refitted for FCM), revival
triggered by the soft mass $N_j$, and the closing step, which now returns the fuzzy means of the last epoch. Like
e-means, the direct transfer starts from $k$ random distinct rows of the data.

\subsection{Two Failures of the Direct Transfer}
\label{sec:failures}
\emph{Revival on soft masses picks the wrong clusters.} Under FCM every point gives every cluster a positive weight, so
the soft mass of an unused cluster rarely falls below a tenth of the mean, and revival rarely fires where it is meant
to. A cluster that covers a small, distant group of points, on the other hand, has a small soft mass by construction.
Revival treats it as dead and moves its centroid onto a dense region, where it adds little, while the points of the
group lose their nearby centroid and $R_m$ rises. KDD Cup 99 contains such groups: tight, distant groups of rare
connection records next to large clumps of duplicate records.

\emph{The starting point determines the local minimum.} On some data, FCM started from random rows ends in a markedly
worse local minimum than FCM started from a k-means solution, and a mini-batch optimizer keeps the solution that its
first epoch selects.

\begin{algorithm}[t]
\caption{Fuzzy e-means (\eFCM); lines 1, 5, 6 and 9 differ from e-means}
\label{alg:efcm}
\begin{algorithmic}[1]
\Require data $X$, number of clusters $k$, fuzzifier $m$, fuzzy epochs $T$
\State $C\gets$ one e-means epoch, with revival, from $k$ random distinct rows \Comment{hard phase}
\State $B,\alpha_0\gets$ e-means laws for $(n,k,T)$; $M\gets T\lceil n/B\rceil$; $N_j\gets 2^{-16}$
\For{$t=0,\dots,M-1$ (batches $\mathcal{B}$, reshuffled per epoch)} \Comment{fuzzy phase}
    \State $\alpha\gets \alpha_0\big(0.1+0.9\cdot\tfrac12(1+\cos(\pi t/M))\big)$
    \State $w_{ij}\gets u_{ij}^m$ by \eqref{eq:u}, for $x_i\in\mathcal{B}$ and all $j$
    \State $n_j\gets\sum_{i\in\mathcal{B}}w_{ij}$; \ $s_j\gets\sum_{i\in\mathcal{B}}w_{ij}x_i$
    \State $N_j'\gets(1-\alpha)N_j+\alpha n_j$; \ $c_j\gets c_j+\frac{\alpha n_j}{N_j'}\big(\frac{s_j}{n_j}-c_j\big)$; \ $N_j\gets N_j'$
    \State in the last epoch: $\bar N_j\gets\bar N_j+n_j$; \ $\bar S_j\gets\bar S_j+s_j$
\EndFor \Comment{no revival in the fuzzy phase}
\State \Return $c_j=\bar S_j/\bar N_j$ \Comment{fuzzy means of the last epoch}
\end{algorithmic}
\end{algorithm}

\subsection{The e-FCM Algorithm}
\eFCM\ (Algorithm~\ref{alg:efcm}) addresses both failures with a hard phase and a fuzzy phase. The hard phase is one
epoch of e-means, with its revival, from $k$ random distinct rows (Faiss implementation \cite{douze2024faiss},
v1.16.0). With hard assignments a small count does mean that few points are nearest to the centroid, so relocation
does what it was designed for, and the epoch places the centroids in a k-means solution. The fuzzy phase then runs $T$ epochs of soft EMA updates from these centroids, with the scaling laws for $T$ epochs, the
cosine schedule and the closing step, and without revival. The hard epoch costs less than a fuzzy one, since it needs
one nearest centroid per point instead of $k$ memberships, and a fuzzy epoch does the work of one AO pass.

The fuzzy phase contains three known updates as special cases: AO-FCM ($B=n$, $\alpha=1$), e-means ($m\to1$) and,
with cumulative statistics $N_j\leftarrow N_j+n_j$, mini-batch FCM with MacQueen--Sculley steps, which we call
MB-FCM (Appendix~\ref{app:relation}). Its centroid update is also a gradient step on the reduced objective. Because
the memberships \eqref{eq:u} minimize $J_m$ for the current centroids, $\partial R_m/\partial D_{ij}=u_{ij}^m/n$, so
the gradient of the batch objective $\sum_{i\in\mathcal{B}}\big(\sum_l D_{il}^{-1/(m-1)}\big)^{1-m}$ with respect to
$c_j$ is $2n_j(c_j-s_j/n_j)$, and line~7 of Algorithm~\ref{alg:efcm} is the step $c_j\gets c_j-\eta_j\nabla_{c_j}$
with $\eta_j=\alpha/(2N_j')$. Since $\alpha n_j/N_j'\in[0,1]$, each update is a convex combination of the old centroid
and the fuzzy mean of the batch, so the centroids stay in the convex hull of the data for any $\alpha_0$.

\section{Experimental Setup}
\label{sec:setup}
\begin{table}[t]
\centering
\caption{Datasets and Fuzzifiers}
\label{tab:data}
\setlength{\tabcolsep}{3.2pt}
\footnotesize
\begin{threeparttable}
\begin{tabular}{@{}lrrllll@{}}
\toprule
Data & $n$ & $d$ & $k$ & $m$ & $m_G$ & $\bar u^\star$ \\
\midrule
\multicolumn{7}{@{}l}{\emph{Standard benchmark}} \\
MNIST & 0.05 & 784 & 10, 100 & 1.15, 1.1 & 1.24 & -- \\
Covertype & 0.57 & 54 & 7, 100 & 1.2, 1.5 & 2.50 & -- \\
KDD99 & 0.48 & 118 & 23, 100 & 2, 1.5 & $\infty$ & -- \\
SIFT1M & 1.00 & 128 & 1024, 16384 & 1.05 & 1.28 & -- \\
DEEP1M & 1.00 & 96 & 1024, 16384 & 1.05 & 1.17 & -- \\
CLIP1M & 1.00 & 512 & 1024, 16384 & 1.05 & 1.18 & -- \\
DINO1M & 1.00 & 1024 & 1024, 16384 & 1.03 & 1.09 & -- \\
\midrule
\multicolumn{7}{@{}l}{\emph{Fuzzy benchmark}} \\
Mix3-sep & 1.00 & 3 & 16 & 1.3 & $\infty$ & 0.95 \\
Mix3-mid & 1.00 & 3 & 16 & 1.4 & $\infty$ & 0.85 \\
Mix3-ovl & 1.00 & 3 & 16 & 1.5 & $\infty$ & 0.70 \\
Mix3 & 1.00 & 3 & 256, 4096 & 1.4 & $\infty$ & 0.85 \\
Mix3-10M & 10.00 & 3 & 256 & 1.4 & $\infty$ & 0.85 \\
Mix3-imb & 1.00 & 3 & 100 & 1.3 & $\infty$ & 0.85 \\
Mix16 & 1.00 & 16 & 256 & 1.1 & $\infty$ & 0.85 \\
Mix128 & 1.00 & 128 & 256 & 1.02 & 4.77 & 0.85 \\
BrainWeb & 1.93 & 3 & 3, 100 & 1.8 & $\infty$ & 0.91 \\
BrainWeb-head & 4.11 & 3 & 9 & 1.6 & $\infty$ & 0.86 \\
EuroSAT & 109.49 & 13 & 10, 100 & 1.5 & $\infty$ & -- \\
\bottomrule
\end{tabular}
\begin{tablenotes}[flushleft]
\footnotesize
\item[] $n$: training points ($10^6$); $d$: dimension; $k$: numbers of clusters; $m$: fuzzifier for each $k$; $m_G$: collapse ceiling ($\infty$: none); $\bar u^\star$: true mean largest membership.
\end{tablenotes}
\end{threeparttable}
\end{table}

\subsection{Benchmarks}
Table~\ref{tab:data} lists the datasets, and Appendix~\ref{app:data} describes them in detail. The standard benchmark
holds three classic sets, MNIST \cite{lecun1998} (handwritten digits), Forest Covertype \cite{blackard1999} and the
10\% KDD Cup 99 set (KDD99) of network connections, clustered at their number of classes and at $k=100$. It also holds
four sets of one million image or descriptor embeddings from the e-means study \cite{bessa2026emeans}, SIFT1M
\cite{jegou2011}, DEEP1M \cite{babenko2016}, CLIP1M \cite{radford2021} and DINO1M \cite{simeoni2025dinov3}, clustered
at $k=1024$ and $k=16384$. With 16,384 clusters, these embedding sets have only 61 training points per cluster.

The fuzzy benchmark consists of low-dimensional data whose groups overlap, such as the intensities of a few MRI
channels or the spectral bands of a satellite image, and of Gaussian mixtures in which we control the overlap. We measure overlap by the true mean largest membership
$\bar u^\star$: the probability, under the mixture, of each point's most likely cluster, averaged over the points; it
is 1 for disjoint clusters. The cluster centres of the mixtures lie in three dimensions. Mix3-sep, Mix3-mid and
Mix3-ovl ($k=16$) differ only in overlap; Mix3 has $k=256$ or $k=4096$ clusters and Mix3-10M ten times as many points;
and Mix3-imb imitates the rare groups of KDD99, with cluster sizes that fall as one over their rank and 10 small,
tight, distant clusters. Mix16 and Mix128 place the $k=256$ mixture in 16 or 128 dimensions and add isotropic noise in
all of them. BrainWeb \cite{collins1998brainweb,kwan1999brainweb} is a simulated MRI scan whose voxels come with the
true fraction of each tissue, a soft ground truth. We cluster its 1.93 million brain voxels (three intensities each)
into $k=3$ tissues and into $k=100$ clusters, and the 4.11 million voxels of the whole head (BrainWeb-head) into its
nine tissues. EuroSAT \cite{helber2019eurosat} holds 109.5 million satellite pixels in 13 spectral bands, which we
cluster at $k=10$ and $k=100$; it has no soft ground truth.

\subsection{Choice of the Fuzzifier}
\label{sec:choosem}
No single fuzzifier suits all these data. The common choice $m=2$ \cite{pal1995} lies above the collapse ceiling $m_G$
of MNIST and of the four embedding sets (Table~\ref{tab:data}), and on Covertype, AO-FCM at $m=2$ keeps only 14\%
($k=7$) and 39\% ($k=100$) of its centroids distinct. We therefore choose $m$ for each configuration by a rule that
keeps FCM meaningful (Appendix~\ref{app:fuzz}). On the standard benchmark we take the largest $m$ on a small grid at
which AO-FCM (100 iterations) keeps as many distinct centroids as at the smallest value of the grid (the
distinct-centroid rule). This gives the fuzziest setting at which FCM still returns $k$ distinct clusters: $m=1.03$ to 1.05 on the embedding sets, 1.1 and 1.15
on MNIST and 1.2 to 2 on Covertype and KDD99. These small values still give soft partitions: averaged over held-out
points, the largest membership is 0.53 to 0.91. On the three-dimensional sets of the fuzzy benchmark this rule does not
limit $m$ (Section~\ref{sec:fuzzifier}). Where the true overlap is known, we instead take the probed $m$ whose mean
largest membership on held-out points is closest to $\bar u^\star$ (the membership-matching rule). This gives $m=1.3$ to 1.5 on the
three-dimensional mixtures, 1.8 on BrainWeb and 1.6 on BrainWeb-head.

\subsection{Compared Methods}
All methods start from the same $k$ random distinct rows for a given seed. We use five seeds on most configurations,
and three on the embedding sets, on Mix3 with $k=4096$, on Mix3-10M and on EuroSAT.

\emph{AO-FCM} runs 300 iterations (200 at $k=16384$). We call AO-FCM after its last iteration \emph{converged AO-FCM},
and its training objectives after 25 and 100 iterations AO@25 and AO@100. Twenty-five is the cap that Faiss and e-means
use for Lloyd's algorithm, and 100 is the default cap of MATLAB's \texttt{fcm}.

\emph{\eFCM} runs its hard epoch and then $T\in\{1,2,3,5,10\}$ fuzzy epochs, which makes $T+1$ passes. Each $T$ is a
separate run, because the laws depend on $T$. \emph{The direct transfer} (random rows, revival on soft masses, no hard
epoch) runs for the same $T$.

\emph{MB-FCM} uses the batches of \eFCM\ and the same random start, with cumulative statistics and a constant step,
and without schedule, revival or closing.

\emph{spFCM and oFCM} split the data into 5 or 20 chunks and run weighted FCM on each chunk until the relative decrease
of the objective falls below $10^{-5}$, for at most 100 iterations; \emph{rseFCM} runs the same solver on a 5\% or
20\% sample.

\emph{Two warm-start hybrids} run 50 AO iterations after a cheap start: from rseFCM on a 10\% sample (rseFCM$\to$AO,
the multistage idea of \cite{cheng1998fast}), and from two epochs of e-means from the Faiss library (e-means$\to$AO).

\subsection{Measurements}
We count cost in passes over the training data. An AO iteration and an epoch of any mini-batch method are one pass
each; for spFCM, oFCM and rseFCM we count the points processed by all inner iterations, divided by $n$. Objectives are
computed exactly, with all $k$ memberships, on every training row and on the held-out rows, and we compare them seed
by seed. A run reaches a target when its training objective is within 0.001\% of it. The speed-up of \eFCM\ is the time
AO-FCM takes to reach the target iteration, divided by the time of the \eFCM\ run with the fewest passes that reaches
the target. All methods share one PyTorch membership kernel and keep the data on the GPU, and times exclude
evaluation. Both benchmarks ran on an NVIDIA RTX PRO 6000 Blackwell GPU. Appendix~\ref{app:impl} describes the
kernel, the runs made on a second GPU and our correctness checks.

\begin{figure*}[t]
\centering
\includegraphics[width=\textwidth]{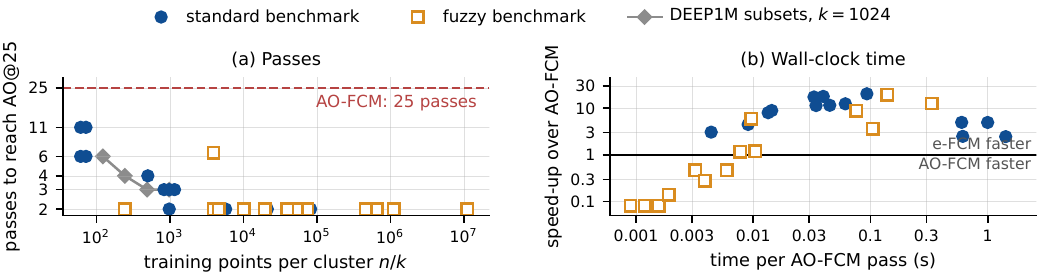}
\caption{\eFCM\ against AO-FCM on all 28 configurations. (a) Passes to reach AO@25, the objective of AO-FCM after 25
iterations (median over seeds; the hard epoch counts as a pass), against training points per cluster. (b) Wall-clock
speed-up to reach AO@25, against the time of one AO-FCM pass.}
\label{fig:summary}
\end{figure*}

\begin{figure*}[t]
\centering
\includegraphics[width=\textwidth]{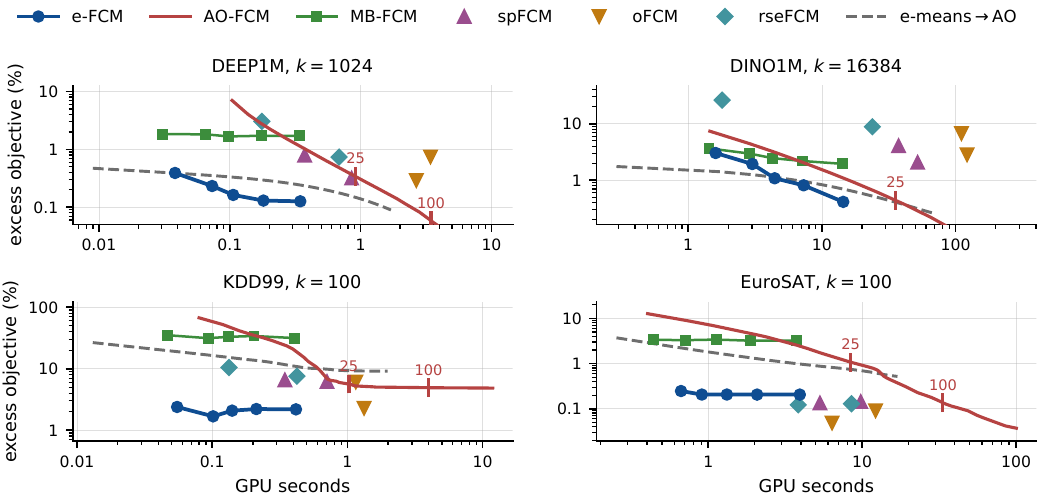}
\caption{Excess training objective over the lowest value found for the same seed by any method, against GPU seconds
(means over seeds). Ticks on AO-FCM: 25 and 100 iterations; other methods: one point per epoch budget, chunk count or
sample size. All configurations: Appendix~\ref{app:curves}.}
\label{fig:curves}
\end{figure*}

\section{Results}
\label{sec:results}
We first ask how many passes \eFCM\ needs to reach the objective of AO-FCM and when fewer passes save time, then where
\eFCM\ ends compared with AO-FCM and the other methods. We then take \eFCM\ apart to see which parts of e-means carry
over, and finally return to the fuzzifier.

\subsection{Passes to Reach AO-FCM}
\label{sec:passes}
A pass over the data is the unit of work that all methods share, and unlike time, its count does not depend on how
fast the hardware is. We
therefore first ask how many passes \eFCM\ needs to reach AO@25. This target is demanding but not final: on the
standard benchmark it lies 0.05 to 3.72\% above converged AO-FCM.

Fig.~\ref{fig:summary}(a) gives the answer for all 28 configurations (Table~\ref{tab:speedfull} lists them). On the
embedding sets with 1,024 clusters (977 points per cluster), \eFCM\ reaches AO@25 in two or three passes. On the
classic sets it needs two to four passes, in every seed except two on Covertype with $k=7$ and one in each KDD99
configuration, which end in a different local minimum (Section~\ref{sec:quality}). On the fuzzy benchmark it needs a
median of two passes on every configuration except Mix128. There AO@25 lies within 0.04\% of converged AO-FCM, and
\eFCM\ ends 0.01 to 0.09\% above it in three of five seeds.

The number of points per cluster decides how many passes \eFCM\ needs. With 16,384 clusters, the embedding sets have
61 points per cluster, and \eFCM\ needs 6 passes on DEEP1M and CLIP1M and 11 on SIFT1M and DINO1M. Subsets of DEEP1M
with $k=1024$ show the trend directly: \eFCM\ needs 6, 4, 3 and 3 passes at 122, 244, 488 and 977 points per cluster.
A plausible reason is that with few points per cluster a batch carries little information about each centroid, so the
moving average needs more batches to settle. With few points per cluster, the training objective also becomes a less
useful target, because AO-FCM starts to fit the training points. With 16,384 clusters, AO@25 lies 0.37 to 0.66\% above
converged AO-FCM on the training points but only 0.08 to 0.18\% above it on held-out points, and on held-out points
\eFCM\ matches AO@25 in three or four passes on SIFT1M, DEEP1M and CLIP1M.

The direct transfer is as fast as \eFCM\ on the embedding sets with 1,024 clusters, but it needs 10 passes with 16,384
clusters and never reaches AO@25 on KDD99 (Table~\ref{tab:speedfull}). Section~\ref{sec:components} traces this
failure to revival.

\subsection{Wall-Clock Time}
\label{sec:time}
Fewer passes save time only if a pass of \eFCM\ costs about as much as a pass of AO-FCM. On the embedding sets it
does: with two or more fuzzy epochs, a fuzzy epoch costs 0.97 to 1.04 times an AO pass, and the hard epoch is cheaper
still (4 to 10\% of a fuzzy epoch with 16,384 clusters). The saving in passes therefore becomes a saving in time
(Fig.~\ref{fig:summary}(b)): \eFCM\ reaches AO@25 11 to 20 times faster than AO-FCM with 1,024 clusters, and 2.5 to 5.0
times faster with 16,384. On the classic sets, where an AO pass takes milliseconds, \eFCM\ is 3.1 to 18 times faster,
and both methods reach AO@25 within about a second. Fig.~\ref{fig:curves} shows the trade-off between objective
and time for four configurations.

A pass of \eFCM\ also carries a fixed cost per batch, for launching GPU kernels and updating the statistics of all
clusters, and this cost matters when a pass of AO-FCM is cheap. The fuzzy benchmark shows where the saving ends.
Where an AO pass takes at least 7.8\,ms (seven configurations of the fuzzy benchmark, including both EuroSAT
settings), \eFCM\ is 1.2 to 20 times faster. Where it takes 1.2 to 5.9\,ms (the other seven, all three-dimensional
with $k\le100$), \eFCM\ is slower, at 0.1 to 0.5 times the speed of AO-FCM. At dimension 16 or below, the Faiss law
picks batches eight times smaller than at higher dimension, so an epoch has 119 to 1,003 batches with a fixed cost of
0.25 to 0.28\,ms each, whereas an AO pass over a million three-dimensional points with $k=16$ takes 1.2\,ms.

\begin{figure*}[t]
\centering
\includegraphics[width=\textwidth]{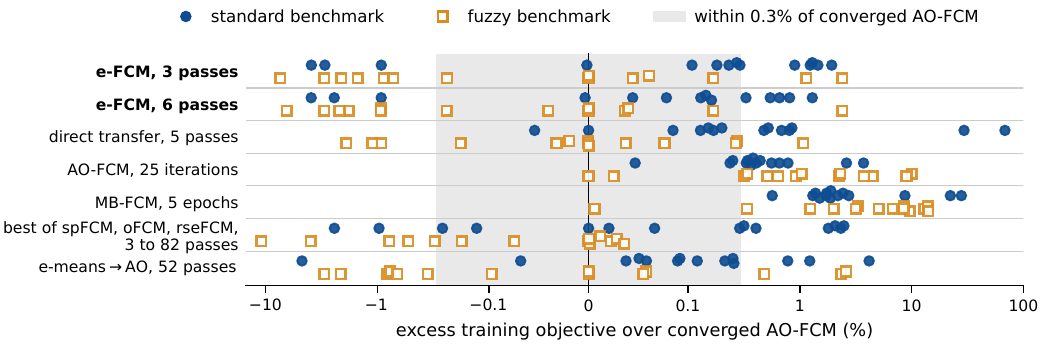}
\caption{Final training objective of each method relative to converged AO-FCM: one marker per configuration (mean
over seeds), standard benchmark in the upper and fuzzy benchmark in the lower half of each row. Values:
Table~\ref{tab:qualityfull}.}
\label{fig:quality}
\end{figure*}

\begin{table*}[t]
\centering
\caption{Ablation of e-FCM after Five Fuzzy Epochs}
\label{tab:ablation}
\setlength{\tabcolsep}{4.6pt}
\footnotesize
\begin{threeparttable}
\begin{tabular}{@{}lrrrrrrrrr@{}}
\toprule
 & \multicolumn{6}{c}{Standard benchmark} & \multicolumn{3}{c}{Fuzzy benchmark} \\
\cmidrule(lr){2-7}\cmidrule(l){8-10}
Change & MNIST & Covertype & KDD99 & KDD99 & DEEP1M & DEEP1M & Mix3 & Mix3-imb & BrainWeb-head \\
 & $k=100$ & $k=100$ & $k=23$ & $k=100$ & $k=1024$ & $k=16384$ & $k=256$ & $k=100$ & $k=9$ \\
\midrule
No hard epoch (random rows) & $\cdot$ & 6.47 & 1.39 & 4.19 & $\cdot$ & 0.12 & $-$0.22 & 2.60 & $-$3.58 \\
Revival on soft masses & $\cdot$ & 0.20 & 46.24 & 26.57 & $\cdot$ & $\cdot$ & $\cdot$ & $\cdot$ & $\cdot$ \\
Cumulative statistics & 0.32 & 0.45 & 1.21 & 3.45 & 0.36 & 0.90 & 2.11 & 2.82 & 0.42 \\
Constant step & $\cdot$ & $\cdot$ & $\cdot$ & $\cdot$ & $\cdot$ & $\cdot$ & $-$0.99 & $-$0.33 & $\cdot$ \\
Batch $\times 1/16$ & $\cdot$ & $\cdot$ & $\cdot$ & $-$0.41 & $\cdot$ & $\cdot$ & $-$2.10 & $-$0.79 & $\cdot$ \\
Peak step $\times 4$ & $\cdot$ & $\cdot$ & $\cdot$ & $\cdot$ & $\cdot$ & 0.19 & $-$1.34 & $-$0.50 & $\cdot$ \\
\bottomrule
\end{tabular}
\begin{tablenotes}[flushleft]
\footnotesize
\item[] Change of the training objective (\%) when one part of e-FCM is changed, mean of 3 seeds; positive values are worse; $\cdot$: below 0.1\% in magnitude.
\end{tablenotes}
\end{threeparttable}
\end{table*}

\subsection{Final Objective}
\label{sec:quality}
Fast early progress would matter less if \eFCM\ then stalled above AO-FCM. Fig.~\ref{fig:quality} compares where each
method ends, relative to converged AO-FCM.

On the embedding sets, \eFCM\ ends close to converged AO-FCM but not below it. After six passes (one hard and five
fuzzy epochs) it is 0.04 to 0.16\%
above with 1,024 clusters, closer than AO@25 (0.24 to 0.41\%). With 16,384 clusters it is 0.33 to 0.81\% above on the
training points, while its held-out objective is below that of converged AO-FCM on three of the four sets.

\eFCM\ ends below converged AO-FCM where AO-FCM from random rows lands in poor minima. After six passes, it is 3.91\% below converged
AO-FCM on Covertype with $k=100$, and 0.92\% ($k=23$) and 2.43\% ($k=100$) below on KDD99 on average, although single
KDD99 seeds range from 6.86\% below to 2.15\% above. On the fuzzy benchmark, \eFCM\ ends more than 0.3\% below
converged AO-FCM on six configurations, within 0.3\% on seven and more than 0.3\% above on one. The largest gain is
on Mix3 with $k=4096$ (6.47\% below), where AO@25 is still 4.50\% above its converged value. On EuroSAT with $k=100$,
\eFCM\ is 0.17\% above on the training pixels but 0.91\% below on held-out pixels.

The k-means start also has a drawback: in some seeds, the hard epoch ends in a k-means solution that is poor for FCM,
and the fuzzy phase cannot leave it. This happens in two of five seeds on Covertype with $k=7$ (7.08\% and 2.57\% above
converged AO-FCM) and in one of five seeds on BrainWeb-head (12.03\% above), which raises the mean excess there to
2.41\%. Which seeds are affected can depend on the GPU (Appendix~\ref{app:impl}) and on the fuzzifier
(Section~\ref{sec:fuzzifier}).

The other methods end higher or need many more passes. MB-FCM, whose per-cluster step shrinks with the accumulated
mass, never reaches AO@25 on the standard benchmark within ten epochs. After five epochs it is 1.31 to 2.74\% above
converged AO-FCM on the embedding sets and up to 27.88\% above on KDD99. On the fuzzy benchmark it is 1.24 to 14.11\%
above, except on Mix128 and on BrainWeb with $k=3$, where AO-FCM itself comes within 0.1\% of its final objective in
at most 14 and 7 iterations.

spFCM, oFCM and rseFCM run weighted FCM to convergence on every chunk or sample. On the standard benchmark with up to
1,024 clusters, the best of the three ends between 2.42\% below and 0.41\% above converged AO-FCM, but its best setting
needs 5 to 77 passes. With 16,384 clusters, a chunk or sample holds only 3 to 12 points per cluster, and the best of
the three ends 1.80 to 2.48\% above converged AO-FCM after 33 to 34 passes (23 to 52\,s), whereas \eFCM\ is 0.3 to
0.8\% above after six passes (3 to 7\,s). On the fuzzy benchmark, the best of the three ends between 10.98\% below and
0.04\% above converged AO-FCM after 3 to 82 passes. The low end is oFCM with 20 chunks on Mix3 with $k=4096$, the lowest
objective of all methods there, but it takes 72.7 passes and 11.07\,s, whereas \eFCM\ reaches 6.47\% below converged
AO-FCM in six passes and 0.563\,s.

The warm-start hybrids end closest to converged AO-FCM on several sets, but at a much higher cost. AO-FCM warm-started
from two epochs of e-means finds the best minimum on Covertype with $k=100$, 4.73\% below converged AO-FCM, and ends
within 0.04 to 0.26\% of it on the embedding sets, closer than \eFCM; its 50 AO iterations, however, take 9.3 to 10
times as long as six-pass \eFCM\ there, and on KDD99 with $k=100$ it ends 4.17\% above converged AO-FCM. AO-FCM
warm-started from a 10\% sample (rseFCM$\to$AO) ends 0.00 to 0.12\% above converged AO-FCM on the embedding sets with
1,024 clusters but 0.34 to 0.52\% above with 16,384.

\begin{figure*}[t]
\centering
\includegraphics[width=\textwidth]{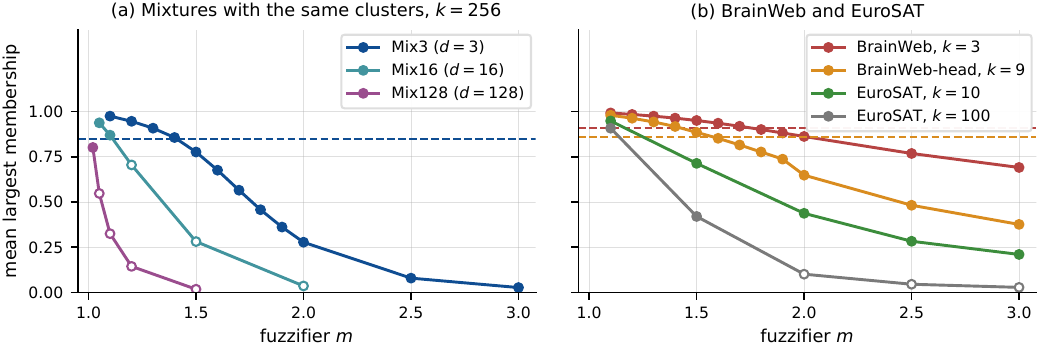}
\caption{Mean largest FCM membership on held-out points after 100 AO-FCM iterations, against the fuzzifier. Dashed:
true value $\bar u^\star$ (Mix16 and Mix128 share that of Mix3); open markers: AO-FCM has merged centroids.}
\label{fig:membership}
\end{figure*}

\subsection{Parts of e-Means That Carry Over}
\label{sec:components}
Table~\ref{tab:ablation} removes or changes one part of \eFCM\ at a time, on six standard configurations that cover
balanced and imbalanced data and both regimes of points per cluster, and on three fuzzy ones; the numbers below are for
five fuzzy epochs unless stated otherwise.

\emph{The moving average is essential.} Replacing it by cumulative statistics, as in MB-FCM, raises the objective on
every configuration, by 0.32 to 3.45\% on the standard benchmark and by 0.42 to 2.82\% on the fuzzy one.

\emph{The hard epoch helps where FCM from random rows lands in poor minima.} Starting the fuzzy phase from random rows
costs 6.47\% on Covertype, 1.39 and 4.19\% on KDD99 and 2.60\% on Mix3-imb, and at most 0.12\% on the other
configurations except BrainWeb-head. There it gains 3.58\%, because one of the three seeds is the one whose hard epoch
picks a poor solution (Section~\ref{sec:quality}).

\begin{table}[t]
\centering
\caption{Revival Triggered by Soft Masses or Hard Counts}
\label{tab:revivalmain}
\setlength{\tabcolsep}{4pt}
\footnotesize
\begin{threeparttable}
\begin{tabular*}{\columnwidth}{@{\extracolsep{\fill}}lrrrrr@{}}
\toprule
 & & \multicolumn{2}{c}{Change (\%)} & \multicolumn{2}{c}{Revivals per run} \\
\cmidrule(lr){3-4}\cmidrule(l){5-6}
Data & $k$ & soft & hard & soft & hard \\
\midrule
KDD99 & 23 & 36.59 & 1.71 & 57 & 3 \\
KDD99 & 100 & 23.11 & 8.83 & 103 & 41 \\
DEEP1M & 16384 & $\cdot$ & $\cdot$ & 189 & 313 \\
DINO1M & 16384 & $-$0.24 & $-$0.25 & 2{,}172 & 2{,}307 \\
\bottomrule
\end{tabular*}
\begin{tablenotes}[flushleft]
\footnotesize
\item[] Change of the training objective (\%) when the fuzzy phase of e-FCM revives clusters (same seed, same hard epoch), after five fuzzy epochs, mean of 3 seeds; positive values are worse; $\cdot$: below 0.1\% in magnitude. Soft: the e-means rule on the moving-average soft mass $\sum_i u_{ij}^m$; hard: on a moving average of the number of batch points nearest to the centroid. On the six configurations of Table~\ref{tab:ablation} not listed, a trigger fires less than once per run and changes the objective by at most 0.03\%. Run on the RTX 6000 Ada.
\end{tablenotes}
\end{threeparttable}
\end{table}

\emph{Revival must be removed from the fuzzy phase.} Adding e-means revival to the fuzzy phase changes nothing on MNIST
and on DEEP1M with 1,024 clusters, where it fires at most once per run, and costs 0.20\% on Covertype. On KDD99 it
costs 46.24\% ($k=23$) and 26.57\% ($k=100$), firing 18 to 139 and 93 to 215 times per run: it moves the centroids of
the rare groups onto the clumps of duplicate rows (Section~\ref{sec:failures}). On the three fuzzy configurations revival
never fires, so Mix3-imb, which imitates the rare groups of KDD99, does not reproduce this failure.

\emph{A trigger on hard counts reduces the damage but does not remove it.} e-means itself triggers revival on a moving
average of hard counts, the number of batch points whose nearest centroid is $c_j$. Under FCM, a cluster that holds a
small group has a small soft mass because the group is small, not because its memberships are soft, so its hard count
may be small as well. Table~\ref{tab:revivalmain} compares the two triggers in runs on our second GPU, against \eFCM\
run there, because the hard epoch can end in a different solution on each GPU (Appendix~\ref{app:impl}). On KDD99 the
hard trigger fires less often than the soft one and cuts the damage from 36.59 to 1.71\% with $k=23$, but only from
23.11 to 8.83\% with $k=100$. Where clusters do die, as on DINO1M with 16,384 clusters, both triggers fire more than
2,000 times per run and lower the objective by 0.24 and 0.25\%. Mass alone, soft or hard, does not distinguish an
empty cluster from a cluster that holds a small, distant group.

\emph{The scaling laws transfer on the standard benchmark, but not in low dimension.} On the standard benchmark, no
tested change of batch size or peak step lowers the objective by more than 0.61\%, except on KDD99 with $k=100$ after
two fuzzy epochs, where a 16 times smaller batch gains 3.37\%. On the fuzzy configurations a 16 times smaller batch
lowers the objective by up to 4.30\% (Mix3, two fuzzy epochs), but smaller batches also add to the fixed cost per batch
that already makes \eFCM\ slow on these sets (Section~\ref{sec:time}).

\emph{The schedule and the closing step barely matter.} On the standard benchmark, a constant step instead of the
cosine schedule changes the objective by at most 0.05\% except on KDD99 with $k=100$, and removing the closing step
by at most 0.10\% (Appendix~\ref{app:ablation}).

\subsection{Fuzzifier at Scale}
\label{sec:fuzzifier}
Three questions about the fuzzifier remain: why the usable values are so small, how the results above depend on $m$,
and whether \eFCM\ can replace the 100 AO-FCM iterations per candidate value that our choice of $m$ used.

\emph{Two effects push the usable $m$ towards 1.} In three dimensions the distinct-centroid rule does not limit $m$:
on every three-dimensional set and on EuroSAT with $k=10$, AO-FCM keeps at least 99\% of its centroids distinct up to
$m=3$, but its memberships become almost uniform long before (Fig.~\ref{fig:membership}). On Mix3 with $k=256$, for
example, the mean largest membership on held-out points falls from 0.78 at $m=1.5$ to 0.28 at $m=2$, while
$\bar u^\star$ is 0.85. The first effect is the number of noise dimensions. Mix16 and Mix128 have the clusters of Mix3,
yet matching gives $m=1.1$ and $m=1.02$, and on Mix128 the mean largest membership stays below $\bar u^\star$ at every
probed value. Noise with standard deviation $\sigma$ in each of $d$ dimensions adds about $d\sigma^2$ to every squared
distance, so the ratios of distances that decide the memberships in \eqref{eq:u} approach 1, and only an $m$ close to 1
keeps the memberships sharp. This suggests that the small fuzzifiers of the embedding sets reflect their dimension
rather than the overlap of their clusters. The second effect is the number of clusters. The membership of a distant
cluster decays as $D^{-1/(m-1)}=r^{-2/(m-1)}$ with its distance $r$, while the number of clusters at distance $r$
grows as $r^{q-1}$ when the centres fill $q$ dimensions; the summed membership of the distant clusters therefore stays
bounded only if $m<1+2/q$, which is 1.67 for $q=3$. In line with this, with 256 or more clusters the mean largest
membership drops sharply between $m=1.5$ and $m=2$ (from 0.75 to 0.14 on Mix3 with $k=4096$), whereas with few
clusters it falls more slowly (from 0.83 to 0.53 on Mix3-mid with $k=16$, and from 0.95 to 0.86 on BrainWeb with
$k=3$). The common choice $m=2$ therefore comes close to the value that matches the true overlap only for
low-dimensional data with few clusters, such as the three tissues of BrainWeb ($m=1.8$).

\emph{\eFCM\ gains most over AO-FCM at the small fuzzifiers of the embedding sets.} Appendix~\ref{app:msweep}
repeats five configurations at other values of $m$. On DEEP1M with 1,024 clusters, AO-FCM converges faster as $m$
grows: AO@25 lies 0.40\% above converged AO-FCM at $m=1.02$ and 0.05\% at $m=1.15$, while \eFCM\ after six passes
stays 0.10 to 0.21\% above; from $m=1.08$ on, AO-FCM itself begins to merge centroids. The effect of the k-means
start, in contrast, persists across the fuzzifier: \eFCM\ ends 5.19 to 8.44\% below converged AO-FCM on Covertype with
$k=100$ at every $m$ up to 1.5, and 0.18 to 8.07\% below on Mix3 with $k=256$ at every $m$ from 1.1 to 2.5 (after
three passes), while on BrainWeb-head the poor hard-epoch solution appears in one or two of three seeds for $m$ from
1.2 to 1.8.

\emph{\eFCM\ can replace AO-FCM for the membership-matching rule, but not for the distinct-centroid rule.} With one
hard epoch shared by all values of $m$ and one, two or five fuzzy epochs per value, \eFCM\ selects the same $m$ as
AO-FCM on eight of the eleven sets with soft truth and a neighbouring value on the other three, which are near ties
(Appendix~\ref{app:mprobe}); a probe over twelve values then costs 25 passes instead of 1,200. In time, the saving
again depends on the cost of a pass: two fuzzy epochs per value of $m$ took 0.18\,s against 13.98\,s for 100 AO
iterations on Mix3 with $k=4096$, but 0.27\,s against 0.13\,s on Mix3-sep. The distinct-centroid rule still needs
AO-FCM, because centroids merge slowly: with five fuzzy epochs, \eFCM\ picks the same $m$ as AO-FCM on only 9 of the
14 standard configurations.

\section{Discussion}
\label{sec:discussion}
\emph{Transfer from k-means to FCM.} The saving in passes carries over from e-means to FCM, because a pass of FCM costs
about as much as a pass of k-means and the moving average works on soft statistics as well as on hard ones. The gain in
final quality does not. In e-means, revival relocates dead clusters, which Lloyd's algorithm cannot revive, so e-means
can end below the loss of Lloyd's algorithm run to convergence \cite{bessa2026emeans}. FCM has no empty clusters, and
its soft memberships smooth the objective, so on the embedding sets AO-FCM from random rows already finds a good
minimum and \eFCM\ stops just above it.

\emph{Two trade-offs in the design.} Removing revival from the fuzzy phase protects small, distant groups, as on KDD99,
but gives up its benefit where clusters do die, as on DINO1M with 16,384 clusters; a relocation rule based on each
cluster's contribution to $R_m$ might serve both. The hard epoch moves FCM out of poor minima, but in some seeds it ends
in a k-means solution that is poor for FCM. Since it costs only 4 to 10\% of a fuzzy epoch on the largest problems,
restarting it a few times and keeping the best fuzzy result is a cheap remedy, which we did not test.

\emph{Choice of method.} For a budget of a few passes, or when the held-out objective matters most, \eFCM\ is the
fastest way to the quality of AO-FCM among the methods we tested. AO-FCM warm-started from e-means comes closest to the
converged training objective on the embedding sets, at about ten times the time of six-pass \eFCM, and where an AO-FCM
pass takes milliseconds, plain AO-FCM is faster.

\emph{Limitations.} All runs use one GPU and dense memberships, and only EuroSAT has more than ten million points;
since the eight largest memberships hold 86 to 100\% of each point's membership on the standard benchmark, truncated
memberships are a natural next step. The e-means laws are used without refitting for FCM, the results rest on three to
five seeds per configuration, and in some seeds the hard epoch ends in a different k-means solution on a different GPU
(Appendix~\ref{app:impl}).

\section{Conclusion}
With a short hard phase that takes over the relocation of centroids, the e-means optimizer makes fuzzy c-means
converge in a few passes: \eFCM\ reaches the objective of 25 AO-FCM iterations in two to four passes when clusters
hold several hundred points or more, with an order of magnitude less GPU time on the one-million-point embedding sets
with 1,024 clusters, and ends close to converged AO-FCM, or below it where FCM from random rows gets stuck. Its gains
shrink when clusters hold few points and turn into a loss on small low-dimensional problems. Next steps are FCM-specific
scaling laws, restarts of the hard phase, a relocation rule based on each cluster's contribution to the objective, and
truncated memberships for very many clusters.

\bibliographystyle{IEEEtran}
\bibliography{refs}

\onecolumn
\appendices

\begin{table*}[!htbp]
\centering
\begin{minipage}[t]{0.46\textwidth}
\centering
\caption{Choice of the Fuzzifier: Standard Benchmark}
\label{tab:fuzzstd}
\setlength{\tabcolsep}{4pt}
\footnotesize
\begin{threeparttable}
\begin{tabular}{@{}lrlll@{}}
\toprule
Data & $k$ & $m$ & Next & $m_e$ \\
\midrule
MNIST & 10 & 1.15 & 1.2 (30\%) & 1.15 \\
 & 100 & 1.1 & 1.15 (70\%) & 1.1 \\
Covertype & 7 & 1.2 & 1.5 (71\%) & 1.5 \\
 & 100 & 1.5 & 2 (39\%) & 1.5 \\
KDD99 & 23 & 2 & -- & 1.2 \\
 & 100 & 1.5 & 2 (89\%) & 1.1 \\
SIFT1M & 1024 & 1.05 & 1.1 (96\%) & 1.1 \\
 & 16384 & 1.05 & 1.1 (91\%) & 1.15/1.1 \\
DEEP1M & 1024 & 1.05 & 1.1 (66\%) & 1.05 \\
 & 16384 & 1.05 & 1.1 (65\%) & 1.1/1.05 \\
CLIP1M & 1024 & 1.05 & 1.1 (56\%) & 1.05 \\
 & 16384 & 1.05 & 1.1 (55\%) & 1.05 \\
DINO1M & 1024 & 1.03 & 1.05 (81\%) & 1.03 \\
 & 16384 & 1.03 & 1.05 (92\%) & 1.05/1.03 \\
\bottomrule
\end{tabular}
\begin{tablenotes}[flushleft]
\footnotesize
\item[] $m$: largest probed value at which AO-FCM (100 iterations) keeps as many distinct centroids as at the smallest probed value. Next: next probed value and the share of centroids that stay distinct there. $m_e$: value chosen by the same rule with e-FCM (two / five fuzzy epochs; one number when both agree).
\end{tablenotes}
\end{threeparttable}
\end{minipage}
\hfill
\begin{minipage}[t]{0.46\textwidth}
\centering
\caption{Choice of the Fuzzifier: Fuzzy Benchmark}
\label{tab:fuzzfuzzy}
\setlength{\tabcolsep}{4pt}
\footnotesize
\begin{threeparttable}
\begin{tabular}{@{}lrcccc@{}}
\toprule
Data & $k$ & $\bar u^\star$ & $m$ & $\bar u_1$ & $m_e$ \\
\midrule
Mix3-sep & 16 & 0.95 & 1.3 & 0.96 & 1.4 \\
Mix3-mid & 16 & 0.85 & 1.4 & 0.88 & 1.5 \\
Mix3-ovl & 16 & 0.70 & 1.5 & 0.73 & 1.5 \\
Mix3 & 256 & 0.85 & 1.4 & 0.86 & 1.4 \\
Mix3 & 4096 & 0.85 & 1.4 & 0.85 & 1.4 \\
Mix3-10M & 256 & 0.85 & 1.4 & 0.86 & 1.4 \\
Mix3-imb & 100 & 0.85 & 1.3 & 0.84 & 1.3 \\
Mix16 & 256 & 0.85 & 1.1 & 0.87 & 1.1 \\
Mix128 & 256 & 0.85 & 1.02 & 0.80 & 1.02 \\
BrainWeb & 3 & 0.91 & 1.8 & 0.90 & 1.8 \\
 & 100 & -- & 1.8 & -- & 1.8 \\
BrainWeb-head & 9 & 0.86 & 1.6 & 0.85 & 1.5 \\
EuroSAT & 10 & -- & 1.5 & -- & 1.5 \\
 & 100 & -- & 1.5 & -- & 1.5 \\
\bottomrule
\end{tabular}
\begin{tablenotes}[flushleft]
\footnotesize
\item[] $\bar u^\star$: true mean largest membership; $m$: probed value whose mean largest FCM membership on held-out points, $\bar u_1$, is closest to $\bar u^\star$; $m_e$: value chosen by the same rule with e-FCM (two fuzzy epochs). EuroSAT takes the distinct-centroid rule at $k=100$ for both $k$; BrainWeb at $k=100$ takes its value at $k=3$.
\end{tablenotes}
\end{threeparttable}
\end{minipage}
\end{table*}

\begin{table*}[t]
\centering
\caption{Passes and Speed-Up of e-FCM per Configuration}
\label{tab:speedfull}
\setlength{\tabcolsep}{5pt}
\footnotesize
\begin{threeparttable}
\begin{tabular}{@{}llrrrrrrrr@{}}
\toprule
 & & & & & \multicolumn{4}{c}{e-FCM} & Direct \\
\cmidrule(lr){6-9}\cmidrule(l){10-10}
 & & & & & \multicolumn{2}{c}{AO@25} & AO@100 & AO@25 held-out & AO@25 \\
\cmidrule(lr){6-7}
Data & $k$ & $m$ & $n/k$ & s/pass & passes & speed-up & passes & passes & passes \\
\midrule
\multicolumn{10}{@{}l}{\emph{Standard benchmark}} \\
MNIST & 10 & 1.15 & 5,000 & 0.0044 & 2 & 3.1 & 3 & 2 & 2 \\
 & 100 & 1.1 & 500 & 0.014 & 4 & 8.9 & 6$^{2/5}$ & 3 & 3 \\
Covertype & 7 & 1.2 & 81,573 & 0.0091 & 2$^{3/5}$ & 4.5 & 2$^{3/5}$ & 2$^{3/5}$ & 2 \\
 & 100 & 1.5 & 5,710 & 0.033 & 2 & 17 & 2 & 2 & 1 \\
KDD99 & 23 & 2 & 21,044 & 0.013 & 2$^{4/5}$ & 8.0 & 2$^{2/5}$ & 2$^{4/5}$ & -- \\
 & 100 & 1.5 & 4,840 & 0.040 & 2$^{4/5}$ & 18 & 2$^{3/5}$ & 2$^{4/5}$ & -- \\
SIFT1M & 1024 & 1.05 & 977 & 0.034 & 3 & 11 & -- & 3 & 3 \\
 & 16384 & 1.05 & 61 & 0.61 & 11 & 2.5 & -- & 3 & 10 \\
DEEP1M & 1024 & 1.05 & 977 & 0.034 & 3 & 12 & -- & 2 & 3 \\
 & 16384 & 1.05 & 61 & 0.60 & 6 & 5.0 & -- & 3 & 10 \\
CLIP1M & 1024 & 1.05 & 977 & 0.061 & 3 & 12 & 4$^{1/3}$ & 2 & 2 \\
 & 16384 & 1.05 & 61 & 1.0 & 6 & 4.9 & -- & 4 & 10 \\
DINO1M & 1024 & 1.03 & 977 & 0.093 & 2 & 20 & 6$^{1/3}$ & 2 & 3 \\
 & 16384 & 1.03 & 61 & 1.4 & 11 & 2.5 & -- & 11 & 10 \\
\midrule
\multicolumn{10}{@{}l}{\emph{Fuzzy benchmark}} \\
Mix3-sep & 16 & 1.3 & 62,500 & 0.0012 & 2 & 0.1 & 3 & 2 & 1 \\
Mix3-mid & 16 & 1.4 & 62,500 & 0.0012 & 2 & 0.1 & 2 & 2$^{4/5}$ & 1 \\
Mix3-ovl & 16 & 1.5 & 62,500 & 0.0012 & 2 & 0.1 & 2 & 2$^{4/5}$ & 1 \\
Mix3 & 256 & 1.4 & 3,906 & 0.0078 & 2$^{4/5}$ & 1.2 & 2$^{4/5}$ & 2$^{4/5}$ & 1 \\
Mix3 & 4096 & 1.4 & 244 & 0.14 & 2 & 20 & 2 & 2 & 2 \\
Mix3-10M & 256 & 1.4 & 39,062 & 0.075 & 2 & 8.9 & 2 & 2 & 1 \\
Mix3-imb & 100 & 1.3 & 10,000 & 0.0032 & 2 & 0.5 & 2 & 2 & 1 \\
Mix16 & 256 & 1.1 & 3,906 & 0.0079 & 2$^{4/5}$ & 1.2 & 2$^{3/5}$ & 2 & 1$^{3/5}$ \\
Mix128 & 256 & 1.02 & 3,906 & 0.0096 & 6.5$^{2/5}$ & 5.9 & 2$^{1/5}$ & 2$^{2/5}$ & 3$^{3/5}$ \\
BrainWeb & 3 & 1.8 & 644,621 & 0.0019 & 2 & 0.1 & 2 & 2 & 1 \\
 & 100 & 1.8 & 19,339 & 0.0059 & 2 & 0.5 & 3 & 2 & 1 \\
BrainWeb-head & 9 & 1.6 & 456,149 & 0.0039 & 2$^{4/5}$ & 0.3 & 2$^{4/5}$ & 2$^{3/5}$ & 1 \\
EuroSAT & 10 & 1.5 & 10,948,608 & 0.10 & 2 & 3.6 & 2 & 2 & 1 \\
 & 100 & 1.5 & 1,094,861 & 0.33 & 2 & 13 & -- & 2 & 1 \\
\bottomrule
\end{tabular}
\begin{tablenotes}[flushleft]
\footnotesize
\item[] Passes: median over the seeds that reached the target, hard epoch included; superscript: share of seeds that reached it; --: no seed did within 11 passes. Speed-up: wall-clock time of AO-FCM to its 25th iteration divided by that of e-FCM. $n/k$: training points per cluster; s/pass: seconds per AO-FCM iteration. Direct: the direct transfer, which has no hard epoch. All times on the RTX PRO 6000 Blackwell.
\end{tablenotes}
\end{threeparttable}
\end{table*}

\section{Relation to Known Methods}
\label{app:relation}
\emph{Special cases.} With $B=n$ and $\alpha=1$, each fuzzy epoch is one AO-FCM iteration. As $m\to1$, the weights
become one-hot and the update becomes that of e-means. With cumulative statistics, $N_j\leftarrow N_j+n_j$, it becomes
mini-batch FCM with MacQueen--Sculley steps (MB-FCM); with batches of one point and a forgetting factor, it is close to
recursive FCM \cite{dovzan2011}. EMA statistics with soft weights are also the fuzzy counterpart of incremental EM with
decayed sufficient statistics \cite{neal1998,cappe2009}. Section~\ref{sec:method} shows that the centroid update is a
gradient step on the reduced objective with a step size $\alpha/(2N_j')$ for each cluster.

\emph{Closing.} The returned centroids are one AO-style update computed from all memberships of the last epoch, which
were evaluated against slowly moving centroids.

\emph{Cost.} A fuzzy epoch does the work of one AO pass: two matrix products per batch ($XC^\top$ for the distances
and $W^\top X$ for the sums) and $O(Bk)$ element-wise work. Like our AO-FCM, which also processes the data in blocks,
it never stores the $n\times k$ membership matrix.

\section{Datasets}
\label{app:data}
\emph{Standard benchmark.} MNIST \cite{lecun1998} pixels are scaled to $[0,1]$. Forest Covertype \cite{blackard1999}
has 54 features, min--max scaled. The 10\% KDD Cup 99 set (KDD99) keeps 38 numeric features through $\log(1+x)$ and
three one-hot categorical features, min--max scaled; about 75\% of its rows are duplicates \cite{tavallaee2009}. These
three classic sets are clustered at their number of classes and at $k=100$, and their held-out points are 10,000
random rows. SIFT1M \cite{jegou2011}, DEEP1M \cite{babenko2016}, CLIP1M (CLIP embeddings of ImageNet images
\cite{radford2021}) and DINO1M (DINOv3 ViT-L features \cite{simeoni2025dinov3}) are the one-million-vector training
sets and held-out splits of the e-means study \cite{bessa2026emeans}, with 10,000 held-out points (20,000 for CLIP1M).

\emph{Gaussian mixtures.} The cluster centres lie in three dimensions, with variance $1/i$ along the $i$-th axis, and
each point is its centre plus isotropic Gaussian noise. For each point we take the probability, under the mixture, of
its most likely cluster; the average over the points is the true mean largest membership $\bar u^\star$. The noise
level is set by bisection so that $\bar u^\star$ takes a target value. Mix3-sep, Mix3-mid and Mix3-ovl ($k=16$) have
$\bar u^\star=0.95$, 0.85 and 0.70. Mix3 with $k=256$ and $k=4096$ has 3,906 and 244 points per cluster, and Mix3-10M
is the $k=256$ mixture with ten times as many points. Mix3-imb has 90 clusters whose sizes fall as one over their rank
(Zipf's law) and 10 small clusters, with 0.02\% of the points each, that lie three times farther out and are three
times tighter. Mix16 and Mix128 place the $k=256$ mixture in a random three-dimensional subspace of 16 or 128
dimensions and add isotropic noise in all of them. The added noise barely changes which cluster a point most likely
came from, so their true memberships are nearly those of Mix3, but it adds about the same amount to every squared
distance.

\emph{BrainWeb} \cite{collins1998brainweb,kwan1999brainweb} is a simulated MRI scan of a normal head: T1-, T2- and
proton-density-weighted images with 1\,mm voxels, 3\% noise and 20\% intensity non-uniformity, together with the
fraction of each tissue in every voxel. These fractions are a soft ground truth, because a voxel on a tissue boundary
contains several tissues. We cluster the 1.93 million voxels that are at least half brain tissue (three intensities
each) into cerebrospinal fluid, grey matter and white matter ($k=3$) and into $k=100$ clusters, and the 4.11 million
voxels of the whole head (BrainWeb-head) into its nine tissues.

\emph{EuroSAT} \cite{helber2019eurosat} consists of 27,000 Sentinel-2 satellite images of $64\times64$ pixels in 13
spectral bands. We cluster its 109.5 million pixels, holding out 1\% of the images, at $k=10$, its number of land-cover
classes, and at $k=100$. Its bands are strongly correlated (the first principal direction holds 65\% of the variance),
so its pixels behave like low-dimensional data. Each set of the fuzzy benchmark has 10,000 held-out points.

\section{Choice of the Fuzzifier}
\label{app:fuzz}
\begin{table*}[t]
\centering
\caption{Final Objective per Configuration}
\label{tab:qualityfull}
\setlength{\tabcolsep}{4.4pt}
\footnotesize
\begin{threeparttable}
\begin{tabular}{@{}llrrrrrrrrrr@{}}
\toprule
 & & & \multicolumn{2}{c}{AO@25} & \multicolumn{3}{c}{e-FCM} & Direct & MB-FCM & Best chunk & e-means \\
\cmidrule(lr){4-5}\cmidrule(lr){6-8}
 & & & & & 3 passes & \multicolumn{2}{c}{6 passes} & 5 passes & 5 passes & or sample & $\to$AO \\
\cmidrule(lr){7-8}
Data & $k$ & $m$ & train & held-out & train & train & held-out & train & train & train & train \\
\midrule
\multicolumn{12}{@{}l}{\emph{Standard benchmark}} \\
MNIST & 10 & 1.15 & 0.05 & 0.03 & 0.00 & 0.00 & 0.00 & $-$0.05 & 0.57 & $-$0.13$^o$ & $-$0.07 \\
 & 100 & 1.1 & 0.25 & 0.23 & 0.27 & 0.14 & 0.15 & 0.13 & 1.38 & 0.32$^o$ & 0.05 \\
Covertype & 7 & 1.2 & 0.38 & 0.42 & 1.29 & 1.29 & 1.25 & 0.00 & 1.51 & 0.00$^o$ & 0.78 \\
 & 100 & 1.5 & 2.62 & 2.78 & $-$3.90 & $-$3.91 & $-$3.28 & 0.85 & 8.72 & $-$0.97$^o$ & $-$4.73 \\
KDD99 & 23 & 2 & 3.72 & 3.29 & $-$0.92 & $-$0.92 & $-$0.87 & 68.39 & 22.22 & $-$0.26$^o$ & 1.23 \\
 & 100 & 1.5 & 0.78 & 0.63 & $-$2.95 & $-$2.43 & $-$2.49 & 29.48 & 27.88 & $-$2.42$^o$ & 4.17 \\
SIFT1M & 1024 & 1.05 & 0.41 & 0.26 & 0.29 & 0.16 & 0.14 & 0.20 & 1.88 & 0.41$^s$ & 0.12 \\
 & 16384 & 1.05 & 0.66 & 0.09 & 1.45 & 0.66 & $-$0.14 & 0.81 & 2.74 & 2.48$^s$ & 0.26 \\
DEEP1M & 1024 & 1.05 & 0.34 & 0.30 & 0.23 & 0.13 & 0.11 & 0.17 & 1.72 & 0.29$^o$ & 0.09 \\
 & 16384 & 1.05 & 0.56 & 0.08 & 1.24 & 0.54 & $-$0.13 & 0.69 & 2.44 & 2.33$^s$ & 0.21 \\
CLIP1M & 1024 & 1.05 & 0.33 & 0.35 & 0.18 & 0.08 & 0.08 & 0.15 & 1.74 & 0.02$^o$ & 0.09 \\
 & 16384 & 1.05 & 0.37 & 0.18 & 0.90 & 0.33 & $-$0.06 & 0.48 & 1.90 & 1.80$^s$ & 0.06 \\
DINO1M & 1024 & 1.03 & 0.24 & 0.28 & 0.11 & 0.04 & 0.05 & 0.08 & 1.31 & 0.07$^o$ & 0.04 \\
 & 16384 & 1.03 & 0.44 & 0.13 & 1.93 & 0.81 & 0.27 & 0.52 & 2.19 & 2.07$^s$ & 0.25 \\
\midrule
\multicolumn{12}{@{}l}{\emph{Fuzzy benchmark}} \\
Mix3-sep & 16 & 1.3 & 10.05 & 10.04 & $-$3.00 & $-$3.00 & $-$2.95 & $-$1.90 & 9.63 & 0.02$^s$ & $-$2.99 \\
Mix3-mid & 16 & 1.4 & 0.64 & 0.63 & $-$0.88 & $-$0.93 & $-$0.89 & $-$0.93 & 6.76 & $-$0.53$^o$ & $-$0.35 \\
Mix3-ovl & 16 & 1.5 & 0.92 & 0.69 & $-$0.24 & $-$0.24 & $-$0.19 & $-$0.02 & 3.30 & $-$0.07$^o$ & $-$0.10 \\
Mix3 & 256 & 1.4 & 2.24 & 2.35 & $-$1.50 & $-$1.81 & $-$1.71 & 0.27 & 13.00 & $-$3.94$^o$ & $-$0.82 \\
Mix3 & 4096 & 1.4 & 4.50 & 4.59 & $-$7.41 & $-$6.47 & $-$5.43 & $-$1.11 & 5.09 & $-$10.98$^o$ & $-$2.13 \\
Mix3-10M & 256 & 1.4 & 3.75 & 3.79 & $-$2.11 & $-$2.18 & $-$2.03 & $-$0.18 & 13.96 & $-$0.31$^o$ & $-$0.77 \\
Mix3-imb & 100 & 1.3 & 2.31 & 2.46 & $-$0.72 & $-$0.94 & $-$1.00 & 1.08 & 8.55 & $-$0.78$^o$ & $-$0.66 \\
Mix16 & 256 & 1.1 & 0.32 & 0.30 & 0.06 & $-$0.04 & $-$0.06 & 0.27 & 1.24 & $-$0.18$^o$ & 0.06 \\
Mix128 & 256 & 1.02 & 0.03 & 0.03 & 0.04 & 0.04 & 0.02 & 0.04 & 0.34 & 0.03$^o$ & 0.05 \\
BrainWeb & 3 & 1.8 & 0.00 & 0.00 & 0.00 & 0.00 & 0.00 & 0.00 & 0.01 & 0.00$^s$ & 0.00 \\
 & 100 & 1.8 & 9.00 & 8.72 & 1.14 & 0.04 & 0.07 & 0.08 & 14.11 & 0.04$^o$ & 2.36 \\
BrainWeb-head & 9 & 1.6 & 0.51 & 0.66 & 2.41 & 2.41 & 2.13 & 0.00 & 8.54 & 0.00$^o$ & 2.59 \\
EuroSAT & 10 & 1.5 & 0.34 & 0.78 & 0.00 & 0.00 & 0.00 & 0.00 & 2.01 & 0.00$^o$ & 0.00 \\
 & 100 & 1.5 & 1.04 & 0.86 & 0.17 & 0.17 & $-$0.91 & $-$0.03 & 3.17 & 0.01$^o$ & 0.48 \\
\bottomrule
\end{tabular}
\begin{tablenotes}[flushleft]
\footnotesize
\item[] Excess over converged AO-FCM (300 iterations; 200 at $k=16384$) in \%, mean over seeds; negative values lie below converged AO-FCM. Best chunk or sample: the best of spFCM ($^s$), oFCM ($^o$) and rseFCM ($^r$), each at its better setting (3 to 82 passes). e-means$\to$AO: 52 passes.
\end{tablenotes}
\end{threeparttable}
\end{table*}
\emph{Distinct-centroid rule (standard benchmark).} We ran AO-FCM for 100 iterations at each value of a small grid of
$m$ and took the largest value at which it keeps as many distinct centroids as at the smallest value of the grid,
within one percentage point. Two centroids count as coincident when they are closer than 1\% of the root-mean-square
distance between points and their nearest centroids. Table~\ref{tab:fuzzstd} lists the result. One grid step higher, 4
to 70\% of the centroids merge. Only KDD99, whose covariance is dominated by one direction ($\lambda_{\max}=0.58$),
keeps all 23 centroids distinct at $m=2$. The chosen values still give soft partitions: averaged over held-out points,
the largest membership is 0.53 to 0.91, and the eight largest memberships hold 86 to 100\% of a point's membership.

\emph{Membership-matching rule (fuzzy benchmark).} Where the true overlap is known, we take the probed $m$ whose mean
largest membership on held-out points, $\bar u_1$, is closest to $\bar u^\star$, among the values at which AO-FCM keeps
its centroids distinct (Table~\ref{tab:fuzzfuzzy}). For this rule we probe twelve values from 1.1 to 3 on the
three-dimensional sets with soft truth, and five values, finer near 1, on Mix16 and Mix128. EuroSAT has no soft truth;
we use the distinct-centroid rule, which binds only at $k=100$ ($m=1.5$), and take its value for both $k$. BrainWeb at
$k=100$ takes its value at $k=3$. On the three-dimensional sets the distinct-centroid rule would pick the largest
probed value, where memberships are almost uniform; Section~\ref{sec:fuzzifier} discusses why the matched values are
small (Fig.~\ref{fig:membership}).

\section{Implementation and Correctness Checks}
\label{app:impl}
All methods share one PyTorch membership kernel: a matrix product for $\|c\|^2-2x^\top c$, a fused
(\texttt{torch.compile}) log-space evaluation of \eqref{eq:u} and of the per-point objective, and a matrix product for
the weighted sums. TF32 is disabled. The classic sets run in FP64 (they are small, and the duplicate rows of KDD99 make
FP32 round-off visible in $R_m$), all other sets in FP32. Data stay on the GPU. Both benchmarks ran on an NVIDIA RTX
PRO 6000 Blackwell (96\,GB, PyTorch 2.14), the fuzzy benchmark with no other job on the GPU. The ablation, the
fuzzifier sweep and most fuzzifier probes on the fuzzy sets, the revival test (Tables~\ref{tab:revivalmain}
and~\ref{tab:revival}) and the
fuzzifier probes with \eFCM\ (Appendix~\ref{app:mprobe}) ran on an NVIDIA RTX 6000 Ada (48\,GB, PyTorch 2.10) while
another user's training job ran on the same GPU. We use these runs only for passes and objectives, which do not depend
on the load, and every ablation compares with \eFCM\ run on the same GPU; the probe times in Appendix~\ref{app:mprobe}
were measured on the Blackwell GPU. Times are wall-clock seconds with GPU synchronization and exclude evaluation.

We checked the kernel against a dense FP64 reference (relative error of the weighted sums at most
$2.4\times10^{-5}$), checked that the fuzzy update with $B=n$, $\alpha=1$ and neither schedule nor closing reproduces
AO-FCM to within $10^{-6}$ (largest difference $8.5\times10^{-7}$), and checked that our implementation with hard
assignments reproduces the Faiss e-means library: the same batch $B=32{,}768$ and step $\alpha_0=0.696$ on DEEP1M with
$k=1024$ and $T=3$, and a test mean squared error of 0.46941 against 0.46940 over three seeds. The first two checks
also hold on the Ada GPU ($2.4\times10^{-5}$ and $9.6\times10^{-7}$). The result of the hard epoch depends on the GPU,
presumably because round-off sends some seeds into a different k-means solution. On Covertype and KDD99, \eFCM\ on the
RTX 6000 Ada differs from \eFCM\ on the Blackwell GPU by up to 4.9\% per seed. An earlier run of the fuzzy benchmark on
the Ada GPU shows how large the effect can be. AO-FCM agrees with the Blackwell run to within 0.03\% after 25
iterations, but after six passes of \eFCM\ one seed of Mix3-sep ends 44.31\%
above converged AO-FCM on the Ada GPU and at converged AO-FCM on the Blackwell GPU, nine other seeds differ by 1.07 to
2.13 percentage points, and on Mix128 \eFCM\ reaches AO@25 in four of five seeds on the Ada GPU but in two on the
Blackwell GPU.

\section{Results per Configuration}
\label{app:full}
\begin{table*}[t]
\centering
\caption{Ablation of e-FCM, All Variants}
\label{tab:ablationfull}
\setlength{\tabcolsep}{4.6pt}
\footnotesize
\begin{threeparttable}
\begin{tabular}{@{}lrrrrrrrrr@{}}
\toprule
 & \multicolumn{6}{c}{Standard benchmark} & \multicolumn{3}{c}{Fuzzy benchmark} \\
\cmidrule(lr){2-7}\cmidrule(l){8-10}
Change & MNIST & Covertype & KDD99 & KDD99 & DEEP1M & DEEP1M & Mix3 & Mix3-imb & BrainWeb-head \\
 & $k=100$ & $k=100$ & $k=23$ & $k=100$ & $k=1024$ & $k=16384$ & $k=256$ & $k=100$ & $k=9$ \\
\midrule
\multicolumn{10}{@{}l}{\emph{After two fuzzy epochs (three passes)}} \\
No hard epoch (random rows) & 0.07 & 6.18 & 1.40 & 3.81 & 0.13 & 0.28 & 0.03 & 2.51 & $-$3.58 \\
Revival on soft masses & 0.00 & 0.20 & 25.15 & 35.85 & 0.00 & 0.11 & 0.00 & 0.00 & 0.00 \\
Direct transfer (both of the above) & 0.07 & 6.96 & 50.80 & 34.48 & 0.13 & 0.40 & 0.07 & 2.70 & $-$3.58 \\
Cumulative statistics & 0.22 & 0.45 & 1.26 & 3.77 & 0.28 & 0.61 & 2.03 & 2.16 & 0.43 \\
Constant step & $-$0.05 & $-$0.01 & 0.00 & $-$0.25 & $-$0.04 & $-$0.03 & $-$0.14 & $-$0.30 & 0.00 \\
No closing step & 0.00 & 0.00 & 0.01 & 0.02 & 0.00 & 0.10 & 0.01 & 0.00 & 0.00 \\
Batch $\times 1/16$ & $-$0.09 & $-$0.61 & $-$0.02 & $-$3.37 & $-$0.06 & 0.21 & $-$4.30 & $-$0.76 & 0.00 \\
Batch $\times 1/4$ & $-$0.07 & $-$0.05 & $-$0.02 & 0.29 & $-$0.04 & 0.15 & 0.08 & $-$0.19 & 0.00 \\
Batch $\times 4$ & 0.13 & 0.25 & 0.03 & 0.88 & 0.17 & 0.05 & 0.16 & 0.36 & 0.21 \\
Peak step $\times 1/4$ & 0.15 & 0.24 & 0.09 & 0.69 & 0.18 & 0.48 & 1.19 & 1.07 & 0.22 \\
Peak step $\times 4$ (capped at 1) & $-$0.08 & 0.00 & 0.00 & $-$0.58 & $-$0.05 & 0.85 & $-$1.13 & $-$1.04 & 0.00 \\
\midrule
\multicolumn{10}{@{}l}{\emph{After five fuzzy epochs (six passes)}} \\
No hard epoch (random rows) & $-$0.02 & 6.47 & 1.39 & 4.19 & 0.04 & 0.12 & $-$0.22 & 2.60 & $-$3.58 \\
Revival on soft masses & 0.00 & 0.20 & 46.24 & 26.57 & 0.00 & 0.02 & 0.00 & 0.00 & 0.00 \\
Direct transfer (both of the above) & $-$0.02 & 6.85 & 57.74 & 31.87 & 0.04 & 0.14 & $-$0.21 & 2.60 & $-$3.58 \\
Cumulative statistics & 0.32 & 0.45 & 1.21 & 3.45 & 0.36 & 0.90 & 2.11 & 2.82 & 0.42 \\
Constant step & $-$0.02 & $-$0.03 & 0.00 & $-$0.02 & $-$0.03 & 0.05 & $-$0.99 & $-$0.33 & 0.00 \\
No closing step & 0.00 & 0.00 & 0.00 & 0.00 & 0.00 & 0.04 & 0.01 & 0.01 & 0.00 \\
Batch $\times 1/16$ & $-$0.09 & $-$0.02 & $-$0.02 & $-$0.41 & $-$0.06 & 0.00 & $-$2.10 & $-$0.79 & 0.00 \\
Batch $\times 1/4$ & $-$0.07 & $-$0.07 & 0.00 & $-$0.36 & $-$0.04 & $-$0.02 & $-$0.31 & $-$0.24 & 0.00 \\
Batch $\times 4$ & 0.20 & 0.25 & 0.01 & 0.58 & 0.22 & 0.24 & 0.13 & 0.29 & 0.17 \\
Peak step $\times 1/4$ & 0.18 & 0.24 & 0.01 & 0.62 & 0.19 & 0.62 & 0.47 & 1.35 & 0.17 \\
Peak step $\times 4$ (capped at 1) & 0.00 & 0.00 & 0.00 & 0.00 & 0.00 & 0.19 & $-$1.34 & $-$0.50 & 0.00 \\
\bottomrule
\end{tabular}
\begin{tablenotes}[flushleft]
\footnotesize
\item[] Change of the training objective (\%) relative to the full method (same seed, same hard epoch), mean of 3 seeds; positive values are worse. A batch change takes its step from the step law at the new batch size.
\end{tablenotes}
\end{threeparttable}
\end{table*}
Table~\ref{tab:speedfull} lists the passes and speed-ups behind Fig.~\ref{fig:summary}, and
Table~\ref{tab:qualityfull} the final objectives behind Fig.~\ref{fig:quality}. On the fuzzy benchmark, the
configurations below converged AO-FCM by more than 0.3\% after six passes of \eFCM\ are Mix3-mid (by 0.93\%), Mix3-imb
(0.94\%), Mix3 with $k=256$ (1.81\%), Mix3-10M (2.18\%), Mix3-sep (3.00\%) and Mix3 with $k=4096$ (6.47\%). On
Covertype with $k=7$, where two of five seeds end in a poor k-means solution (Section~\ref{sec:quality}), a third seed
ends 3.17\% below converged AO-FCM. The direct transfer, which starts from random rows, avoids the poor seed on
BrainWeb-head; AO-FCM warm-started from two hard epochs does not. AO-FCM warm-started from a 10\% sample
(rseFCM$\to$AO), which Table~\ref{tab:qualityfull} omits, ends between 0.06\% below and 1.41\% above converged AO-FCM
on the classic sets and between 1.00\% below and 1.08\% above on the fuzzy benchmark; Section~\ref{sec:quality} gives
the embedding sets.

\section{Full Ablation}
\label{app:ablation}
\begin{table*}[t]
\centering
\caption{Revival Triggered by Soft Masses or Hard Counts, All Runs}
\label{tab:revival}
\setlength{\tabcolsep}{5pt}
\footnotesize
\begin{threeparttable}
\begin{tabular}{@{}llrrrrrr@{}}
\toprule
 & & & \multicolumn{2}{c}{After the hard epoch} & \multicolumn{3}{c}{From random rows} \\
\cmidrule(lr){4-5}\cmidrule(l){6-8}
Data & $k$ & $m$ & soft & hard & none & soft & hard \\
\midrule
MNIST & 100 & 1.1 & 0.00/0.00 (0) & 0.00/0.00 (0) & 0.07/$-$0.02 & 0.07/$-$0.02 (0) & 0.07/$-$0.02 (0) \\
Covertype & 100 & 1.5 & 0.29/0.00 (0.3) & 0.00/0.00 (0) & 6.75/5.62 & 6.75/6.59 (0.2) & 7.57/7.39 (4) \\
KDD99 & 23 & 2 & 38.89/36.59 (54) & 1.14/1.71 (3) & 1.89/1.88 & 53.98/55.12 (79) & 0.06/2.11 (7) \\
KDD99 & 100 & 1.5 & 24.53/23.11 (126) & 12.25/8.83 (50) & 4.54/4.71 & 37.02/29.17 (154) & 14.25/11.77 (68) \\
DEEP1M & 1024 & 1.05 & 0.00/0.00 (0) & 0.00/0.00 (0) & 0.14/0.06 & 0.14/0.06 (0.2) & 0.14/0.06 (0.3) \\
DEEP1M & 16384 & 1.05 & 0.12/0.02 (377) & 0.21/0.03 (641) & 0.28/0.13 & 0.41/0.14 (530) & 0.51/0.15 (855) \\
DINO1M & 16384 & 1.03 & $-$0.31/$-$0.24 (3036) & $-$0.31/$-$0.25 (3242) & 0.05/0.09 & $-$0.29/$-$0.26 (3152) & $-$0.28/$-$0.26 (3407) \\
Mix3 & 256 & 1.4 & 0.00/0.00 (0) & 0.00/0.00 (0) & 0.03/$-$0.22 & 0.07/$-$0.21 (0.5) & 0.07/$-$0.20 (0.5) \\
Mix3-imb & 100 & 1.3 & 0.00/0.00 (0) & 0.00/$-$0.03 (0.2) & 2.51/2.60 & 2.70/2.60 (0.2) & 2.70/2.60 (0.2) \\
BrainWeb-head & 9 & 1.6 & 0.00/0.00 (0) & 0.00/0.00 (0) & $-$3.58/$-$3.58 & $-$3.58/$-$3.58 (0) & $-$3.58/$-$3.58 (0) \\
\bottomrule
\end{tabular}
\begin{tablenotes}[flushleft]
\footnotesize
\item[] Change of the training objective (\%) relative to e-FCM, which revives no cluster in its fuzzy phase (same seed, same GPU), after two / five fuzzy epochs, mean of 3 seeds; in brackets, revivals per run. Soft: the e-means rule on the moving-average soft mass $\sum_i u_{ij}^m$; hard: the same rule on a moving average of the number of batch points nearest to the centroid; none: random rows without revival. Run on the RTX 6000 Ada.
\end{tablenotes}
\end{threeparttable}
\end{table*}
Table~\ref{tab:ablationfull} gives every variant after two and after five fuzzy epochs. On the standard benchmark, no
tested change of batch size or peak step lowers the objective by more than 0.61\%, except on KDD99 with $k=100$ after
two fuzzy epochs, where a 16 times smaller batch gains 3.37\%; the other changes cost up to 0.88\%. On the fuzzy
configurations the laws are further from the best tested setting. A 16 times smaller batch lowers the objective by
4.30\% on Mix3 after two fuzzy epochs (2.10\% after five) and by 0.76 to 0.79\% on Mix3-imb, and a four times larger
peak step lowers it by 0.50 to 1.34\% on both. Smaller batches, however, mean more batches per epoch and more of the
fixed cost per batch that already makes \eFCM\ slow on these sets (Section~\ref{sec:time}).

On the standard benchmark, replacing the cosine schedule by a constant step changes the objective by at most 0.05\%,
except on KDD99 with $k=100$, where the constant step is 0.25\% better after two fuzzy epochs; on Mix3 it is up to
0.99\% better. Removing the closing step costs at most 0.10\%. In e-means both parts help by a fraction of a percent
\cite{bessa2026emeans}. A plausible reason for the even smaller effect here is that soft memberships already average
each centroid update over many points, which damps the noise that the schedule and the closing step remove in the hard
case.

Revival on soft masses fires at most once per run on MNIST, Covertype and DEEP1M with 1,024 clusters. On DEEP1M with
16,384 clusters it fires 177 to 582 times per run and costs 0.02 to 0.11\%. On KDD99 it costs 25.15 to 46.24\% over
both budgets. The direct transfer combines the random start with revival and inherits both effects.

\section{Revival Triggered by Hard Counts}
\label{app:revival}
\begin{figure*}[t]
\centering
\includegraphics[width=\textwidth]{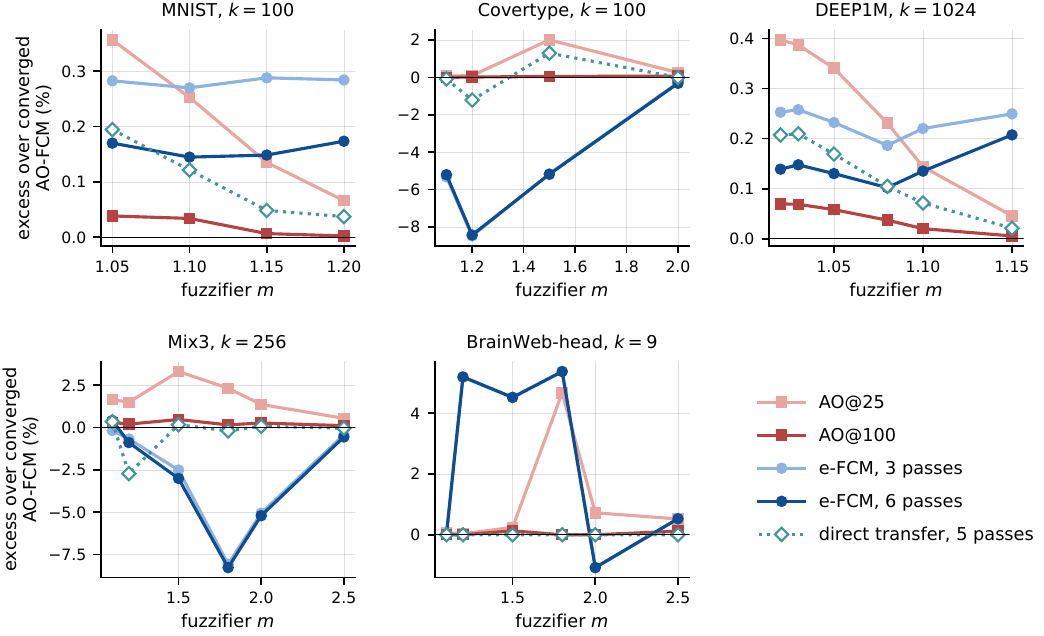}
\caption{Excess training objective over converged AO-FCM against the fuzzifier (means of 3 seeds); negative values
lie below converged AO-FCM. The fuzzifiers used elsewhere are $m=1.1$ (MNIST), 1.5 (Covertype), 1.05 (DEEP1M), 1.4
(Mix3) and 1.6 (BrainWeb-head).}
\label{fig:msweep}
\end{figure*}
Table~\ref{tab:revival} extends Table~\ref{tab:revivalmain} to all ten configurations of the revival test, to two
fuzzy epochs as well as five, and to starts from random rows, as in the direct transfer. Like
Table~\ref{tab:revivalmain}, it compares with \eFCM\ run on the same GPU, because the result of the hard epoch differs
between our two GPUs (Appendix~\ref{app:impl}).

On KDD99 with $k=23$ the hard trigger removes most of the damage. After the hard epoch the objective rises by 1.14 to
1.71\% instead of 36.59 to 38.89\%, and from random rows by 0.06 to 2.11\% instead of 53.98 to 55.12\%. With $k=100$ it
removes only part of the damage (8.83 to 12.25\% instead of 23.11 to 24.53\% after the hard epoch); there it fires 50
times per run instead of 3 with $k=23$, presumably because with 100 clusters more of them cover small groups of rare
records.

Where clusters do die, the hard trigger keeps the benefit of revival. On DINO1M with 16,384 clusters and $m=1.03$, the
setting closest to k-means in our study, both triggers lower the objective by 0.24 to 0.31\%; on DEEP1M with 16,384
clusters the hard trigger costs up to 0.21\%, against 0.12\% for the soft one. On the other configurations the hard
trigger fires at most a few times per run and changes the objective by at most 0.03\% after the hard epoch, but from
random rows on Covertype it raises the objective further (7.39 to 7.57\% instead of 6.59 to 6.75\%). A rule based on
each cluster's contribution to $R_m$ could serve both cases; we leave it to future work.

\section{Dependence on the Fuzzifier}
\label{app:msweep}
Fig.~\ref{fig:msweep} shows how the final objective depends on the fuzzifier; Section~\ref{sec:fuzzifier} summarizes
it. MNIST behaves like DEEP1M: AO@25 lies 0.36\% above converged AO-FCM at $m=1.05$ and 0.07\% at $m=1.2$, while
\eFCM\ after six passes stays at 0.14 to 0.17\%. On Covertype at $m=2$, AO-FCM keeps only 34\% of its centroids
distinct, and \eFCM\ ends 0.32\% below it. On Mix3 with $k=256$, \eFCM\ gains most at $m=1.5$ to 2, where AO@25 is
still 1.37 to 3.31\% above its converged value. On BrainWeb-head, the seeds whose hard epoch picks a poor solution end
8.06 to 15.59\% above converged AO-FCM, while the direct transfer, which starts from random rows, matches converged
AO-FCM at every $m$.

\section{Choosing the Fuzzifier with e-FCM}
\label{app:mprobe}
We reran both rules of Appendix~\ref{app:fuzz} with \eFCM: one hard epoch shared by all values of $m$, then one, two or
five fuzzy epochs per value (column $m_e$ of Tables~\ref{tab:fuzzstd} and~\ref{tab:fuzzfuzzy}). For matching
memberships, \eFCM\ is accurate. At the $m$ selected with AO-FCM, its mean largest membership is within 0.012 of that
of AO-FCM on all eleven sets with soft truth, and the value of $m$ at which it meets $\bar u^\star$, interpolated
linearly between probed values, differs from that of AO-FCM by at most 0.032 on the ten sets where both cross
$\bar u^\star$. It selects the same $m$ on eight sets and a neighbouring value on the other three, which are near ties.
On Mix3-sep and Mix3-mid, the two neighbouring values put the mean largest membership of AO-FCM almost equally far from
$\bar u^\star$ (0.010 and 0.011; 0.024 and 0.026); on BrainWeb-head, the value of \eFCM\ at $m=1.6$ lies 0.018 from
$\bar u^\star$, against 0.006 for AO-FCM. The times in Section~\ref{sec:fuzzifier} were measured on the Blackwell GPU
and exclude the shared hard epoch, which took 0.046\,s on Mix3-sep and 0.037\,s on Mix3 with $k=4096$.

The distinct-centroid rule needs AO-FCM. With five fuzzy epochs, \eFCM\ picks the same $m$ as AO-FCM on 9 of the 14
standard configurations, and with two on 7. It picks one grid step higher on SIFT1M and on Covertype with $k=7$, where
AO-FCM has already merged centroids that \eFCM\ still keeps apart. On KDD99, whose duplicate rows can hold several
centroids, the share of distinct centroids does not fall steadily with $m$ (with $k=23$, AO-FCM keeps 91.3\% at
$m=1.5$ but all at $m=2$), and \eFCM\ picks smaller values.

\section{Objective Against Time on All Configurations}
\label{app:curves}
\begin{figure*}[!htbp]
\centering
\includegraphics[width=\textwidth]{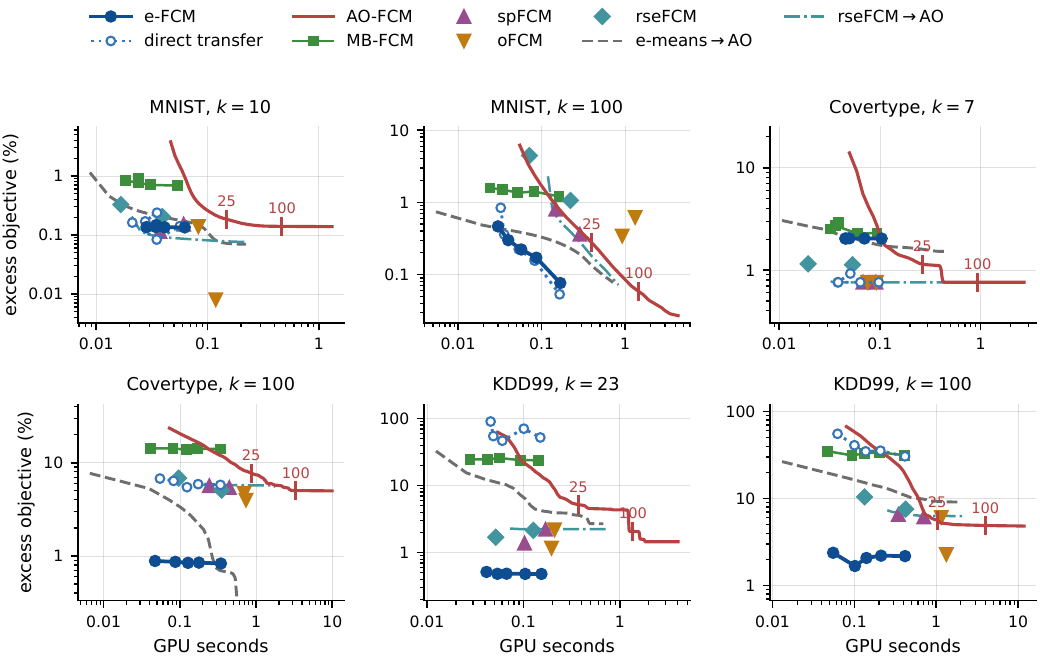}
\caption{Classic sets of the standard benchmark: excess training objective over the lowest value found for the same
seed by any method, against GPU seconds (means over seeds).}
\label{fig:curvesclassic}
\end{figure*}

\begin{figure*}[!htbp]
\centering
\includegraphics[width=\textwidth]{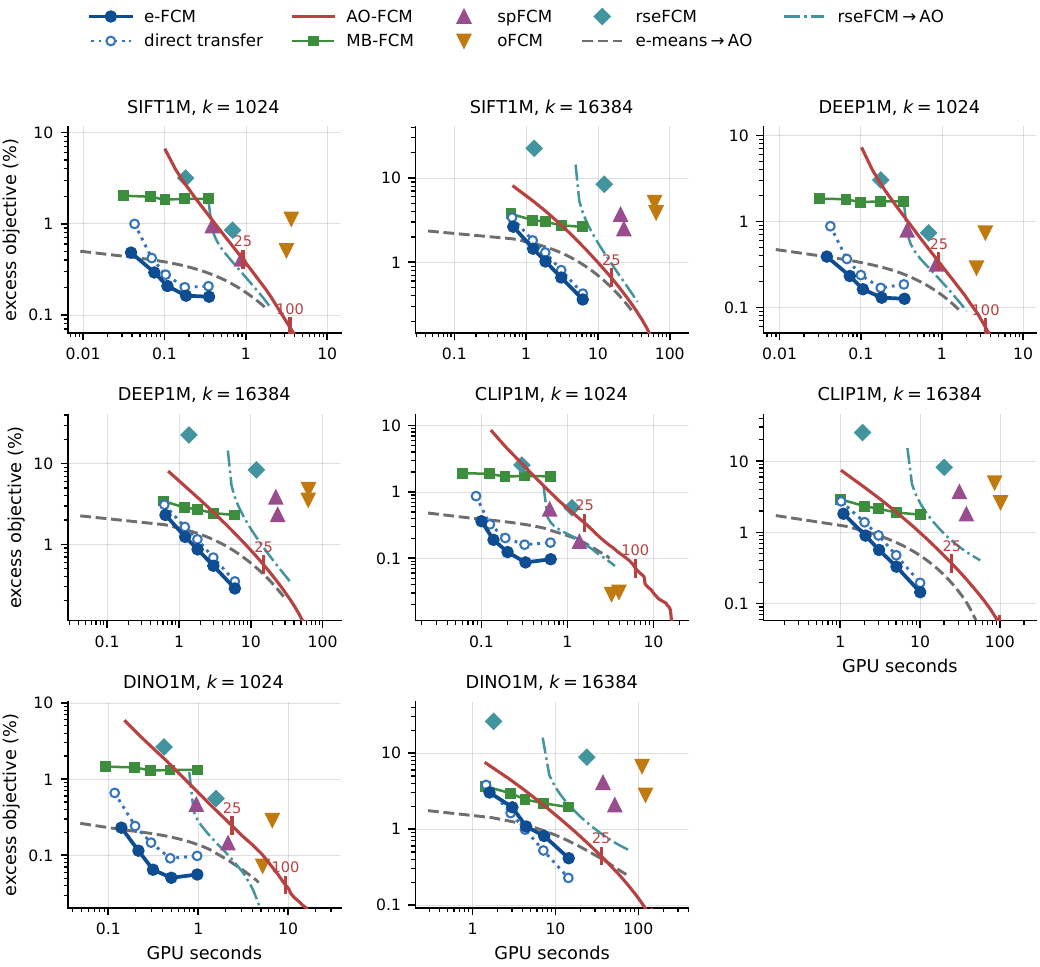}
\caption{Embedding sets of the standard benchmark, as in Fig.~\ref{fig:curvesclassic}.}
\label{fig:curvesemb}
\end{figure*}

\begin{figure*}[!htbp]
\centering
\includegraphics[width=\textwidth]{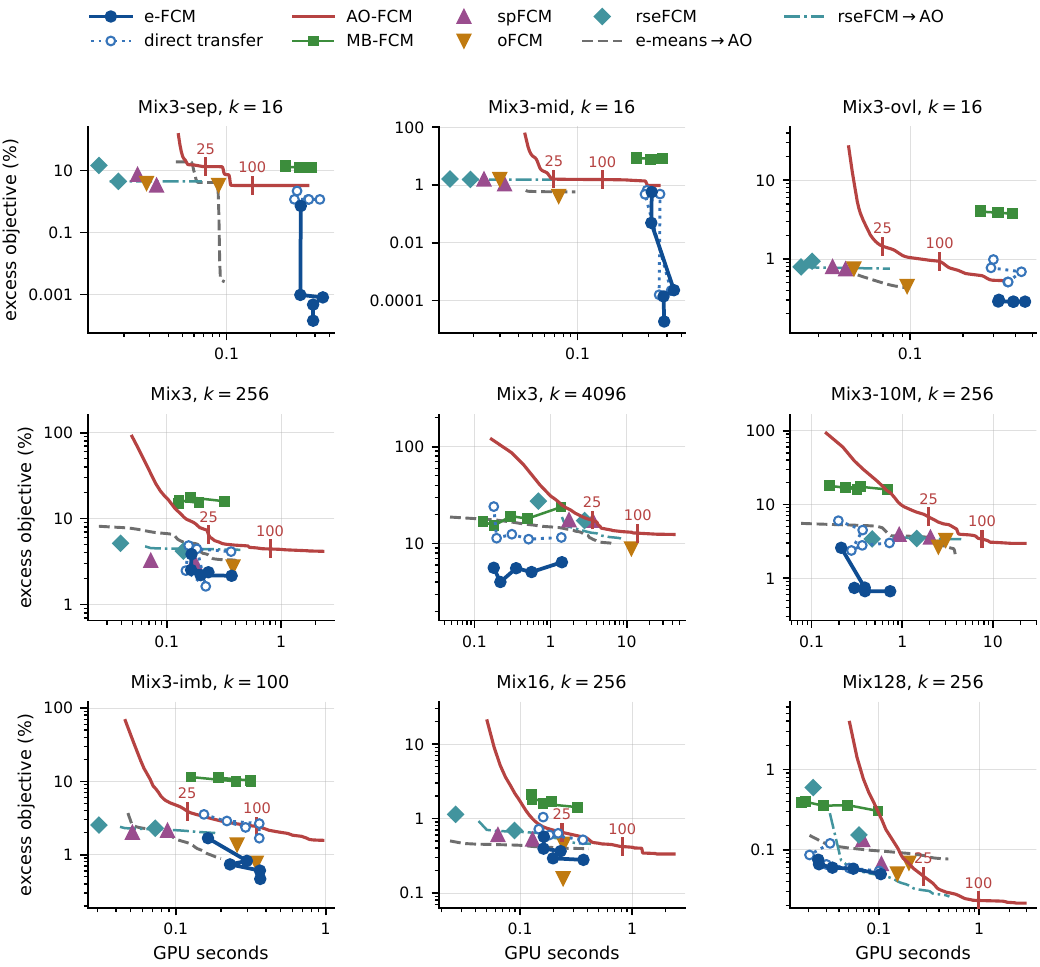}
\caption{Gaussian mixtures of the fuzzy benchmark, as in Fig.~\ref{fig:curvesclassic}.}
\label{fig:curvesmix}
\end{figure*}

\begin{figure*}[!htbp]
\centering
\includegraphics[width=\textwidth]{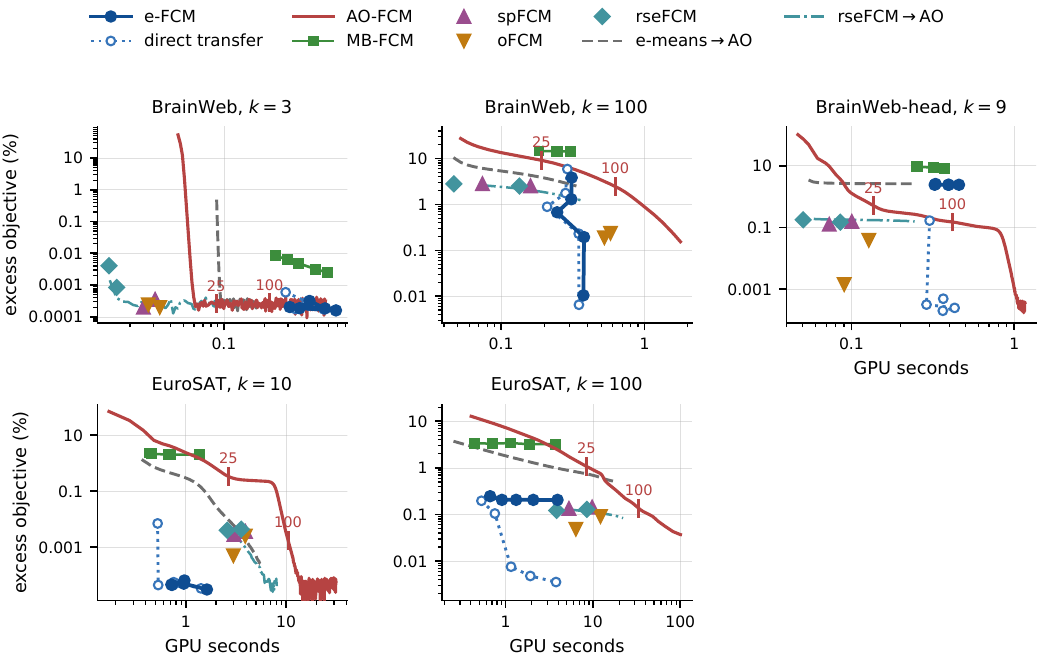}
\caption{BrainWeb and EuroSAT, as in Fig.~\ref{fig:curvesmix}.}
\label{fig:curvesreal}
\end{figure*}

Figs.~\ref{fig:curvesclassic} to~\ref{fig:curvesreal} extend Fig.~\ref{fig:curves} to every configuration and add
the direct transfer and rseFCM$\to$AO. AO-FCM is a curve with ticks at 25 and 100 iterations; \eFCM, the direct transfer
and MB-FCM show one point per epoch budget; spFCM, oFCM and rseFCM one point per chunk count or sample size; and the
two warm-start hybrids are curves over their 50 AO iterations. A curve that reaches the lowest value of its seed leaves
the panel through the bottom.


\end{document}